\pdfoutput=1

\documentclass[11pt]{article}

\usepackage[final]{acl}

\usepackage{times}
\usepackage{latexsym}

\usepackage[T1]{fontenc}

\usepackage[utf8]{inputenc}

\usepackage{microtype}

\usepackage{inconsolata}

\usepackage{microtype}
\usepackage{nicefrac}
\usepackage{url}
\usepackage{hyperref}
\usepackage{booktabs}
\usepackage{enumitem}
\usepackage{color}
\usepackage{colortbl}
\usepackage{float}
\definecolor{mygray}{gray}{.9}
\usepackage{bbding}
\usepackage{amsmath}      
\usepackage{xspace}

\usepackage{enumitem}

\DeclareMathOperator*{\argmax}{arg\,max}

\usepackage{xcolor}
\usepackage{color}

\usepackage[title]{appendix}
\usepackage{algorithm}
\usepackage{algpseudocode}
\usepackage{mathtools}

\usepackage{algpseudocode}
\algrenewcommand\algorithmicrequire{\textbf{Input:}}
\algrenewcommand\algorithmicensure{\textbf{Output:}}
\usepackage{pdfpages}
\usepackage{caption}
\usepackage{subcaption}
\usepackage{footnote}
\usepackage{multirow}
\usepackage{array}

\usepackage[normalem]{ulem}
\usepackage{soul}
\usepackage{wrapfig}

\usepackage{amssymb}
\usepackage{graphicx}
\usepackage{amsmath}

\usepackage{amsmath}
\usepackage{amssymb} 
\usepackage{mathtools}
\usepackage{tikz} 

\usepackage{algorithm}
\usepackage{algpseudocode}

\usepackage{kantlipsum}
\usepackage[breakable, theorems, skins]{tcolorbox}
\tcbset{enhanced}

\DeclareRobustCommand{\mybox}[2][gray!20]{%
\begin{tcolorbox}[   
        breakable,
        left=0pt,
        right=0pt,
        top=0pt,
        bottom=0pt,
        colback=#1,
        colframe=#1,
        width=\dimexpr0.39\textwidth\relax, 
        enlarge left by=0mm,
        boxsep=5pt,
        arc=0pt,outer arc=0pt,
        ]
        #2
\end{tcolorbox}
}

\makeatletter
\def\slashedarrowfill@#1#2#3#4#5{%
  $\m@th\thickmuskip0mu\medmuskip\thickmuskip\thinmuskip\thickmuskip
   \relax#5#1\mkern-7mu%
   \cleaders\hbox{$#5\mkern-2mu#2\mkern-2mu$}\hfill
   \mathclap{#3}\mathclap{#2}%
   \cleaders\hbox{$#5\mkern-2mu#2\mkern-2mu$}\hfill
   \mkern-7mu#4$%
}

\def\rightslashedarrowfilla@{%
  \slashedarrowfill@\relbar\relbar{\raisebox{1.2pt}{$\scriptscriptstyle\diagup$}}\rightarrow}
\newcommand\xslashedrightarrowa[2][]{%
  \ext@arrow 0055{\rightslashedarrowfilla@}{#1}{#2}}

\def\rightslashedarrowfillb@{%
  \slashedarrowfill@\relbar\relbar/\rightarrow}
\newcommand\xslashedrightarrowb[2][]{%
  \ext@arrow 0055{\rightslashedarrowfillb@}{#1}{#2}}

\def\rightslashedarrowfillc@{%
  \slashedarrowfill@\relbar\relbar{\raisebox{.12em}{\tiny/}}\rightarrow}
\newcommand\xslashedrightarrowc[2][]{%
  \ext@arrow 0055{\rightslashedarrowfillc@}{#1}{#2}}

\pgfdeclareshape{slash underlined}
{
  \inheritsavedanchors[from=rectangle] 
  \inheritanchorborder[from=rectangle]
  \inheritanchor[from=rectangle]{north}
  \inheritanchor[from=rectangle]{north west}
  \inheritanchor[from=rectangle]{north east}
  \inheritanchor[from=rectangle]{center}
  \inheritanchor[from=rectangle]{west}
  \inheritanchor[from=rectangle]{east}
  \inheritanchor[from=rectangle]{mid}
  \inheritanchor[from=rectangle]{mid west}
  \inheritanchor[from=rectangle]{mid east}
  \inheritanchor[from=rectangle]{base}
  \inheritanchor[from=rectangle]{base west}
  \inheritanchor[from=rectangle]{base east}
  \inheritanchor[from=rectangle]{south}
  \inheritanchor[from=rectangle]{south west}
  \inheritanchor[from=rectangle]{south east}
  \inheritanchorborder[from=rectangle]
  \foregroundpath{
    \southwest \pgf@xa=\pgf@x \pgf@ya=\pgf@y
    \northeast \pgf@xb=\pgf@x \pgf@yb=\pgf@y
    \pgf@xc=\pgf@xa
    \advance\pgf@xc by .5\pgf@xb
    \pgf@yc=\pgf@ya
    \advance\pgf@xc by -1.3pt
    \advance\pgf@yc by -1.8pt
    \pgfpathmoveto{\pgfqpoint{\pgf@xc}{\pgf@yc}}
    \advance\pgf@xc by  2.6pt
    \advance\pgf@yc by  3.6pt
    \pgfpathlineto{\pgfqpoint{\pgf@xc}{\pgf@yc}}
    \pgfpathmoveto{\pgfqpoint{\pgf@xa}{\pgf@ya}}
    \pgfpathlineto{\pgfqpoint{\pgf@xb}{\pgf@ya}}
 }
}
\tikzset{nomorepostaction/.code=\let\tikz@postactions\pgfutil@empty}

\makeatother
\usepackage{bm} 

\usepackage{amsmath,amssymb}
\usepackage{xspace}
\newcommand{\shorttitle}[0]{\texorpdfstring{\texttt{\textsc{$\delta$-CAUSAL}}}{delta-Causal}\xspace}
\newcommand{\tbfspace}[0]{\;\xspace}

\newcommand{\causalstrength}[2]{\mathcal{CS}(#1 \to #2)}

\usepackage{wrapfig}
\usepackage{longtable}
\makeatletter
\let\oldlt\longtable
\let\endoldlt\endlongtable
\def\longtable{\@ifnextchar[\longtable@i \longtable@ii}
\def\longtable@i[#1]{\begin{figure}[t]
\onecolumn
\begin{minipage}{0.5\textwidth}
\oldlt[#1]
}
\def\longtable@ii{\begin{figure}[t]
\onecolumn
\begin{minipage}{0.5\textwidth}
\oldlt
}
\def\endlongtable{\endoldlt
\end{minipage}
\twocolumn
\end{figure}}
\makeatother

\usepackage{amssymb} 

\newcommand{\RNum}[1]{\texorpdfstring{\uppercase\expandafter{\romannumeral #1}}{\romannumeral #1}}

\newcommand{\icon}[2][0.4cm]{\begin{minipage}{#1}\includegraphics[width=#1]{#2}\end{minipage}}
\newcommand{\bfull}[0]{\icon{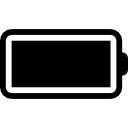}\xspace}
\newcommand{\bempty}[0]{\icon{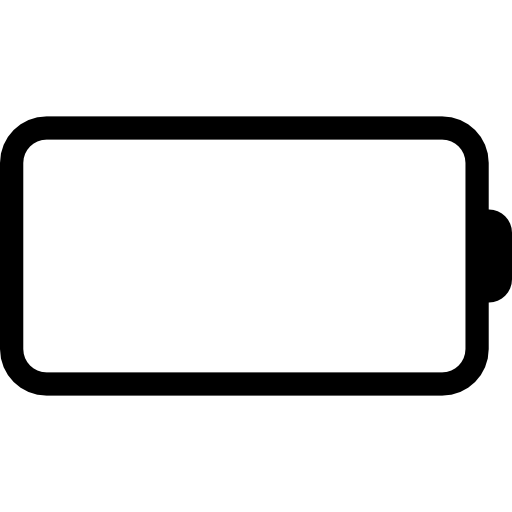}\xspace}

\usepackage{graphicx} 
\usepackage{float} 
\usepackage{adjustbox}

\title{Exploring Defeasibility in Causal Reasoning}

\author {\textbf{Shaobo Cui, Lazar Milikic, Yiyang Feng, Mete Ismayilzada, } \\ 
\textbf{Debjit Paul, Antoine Bosselut, Boi Faltings} \\
        EPFL, Switzerland \\
        \texttt{\{firstname.lastname\}@epfl.ch}
        }

\begin{document}
\maketitle

\begin{abstract}
Defeasibility in causal reasoning implies that the causal relationship between cause and effect can be strengthened or weakened. Namely, the causal strength between cause and effect should increase or decrease with the incorporation of strengthening arguments~(supporters) or weakening arguments~(defeaters), respectively. However, existing works ignore defeasibility in causal reasoning and fail to evaluate existing causal strength metrics in defeasible settings. In this work, we present \shorttitle, the first benchmark dataset for studying defeasibility in causal reasoning. \shorttitle includes around 11K events spanning ten domains, featuring defeasible causality pairs, namely, cause-effect pairs accompanied by supporters and defeaters. We further show that current causal strength metrics fail to reflect the change of causal strength with the incorporation of supporters or defeaters in \shorttitle. 
To this end, we propose \texttt{CESAR}~(\ul{C}ausal \ul{E}mbedding a\ul{S}sociation with \ul{A}ttention \ul{R}ating), a metric that measures causal strength based on token-level causal relationships.  \texttt{CESAR} achieves a significant 69.7\% relative improvement over existing metrics, increasing from 47.2\% to 80.1\% in capturing the causal strength change brought by supporters and defeaters. We further demonstrate even Large Language Models~(LLMs) like GPT-3.5 still lag 4.5 and 10.7 points behind humans in generating supporters and defeaters, emphasizing the challenge posed by \shorttitle.
\end{abstract}
\section{Introduction}
Causality~\cite{pearl2009causality,pearl2018book}, a fundamental concept of artificial intelligence, describes the relationship between two events where one event, namely the \emph{cause}, results in the occurrence of another event, namely the \emph{effect}. Understanding causality enhances decision-making in various areas such as medicine~\cite{kuipers1984causal,michoel2022causal}, disease treatment~\cite{rizzi1994causal,halloran1995causal}, finance~\cite{koonce2011causal,tiffin2019machine}, and law~\cite{hart1985causation,liu2021everything}. 
\begin{figure}
\centering
\includegraphics[width=0.99\linewidth]{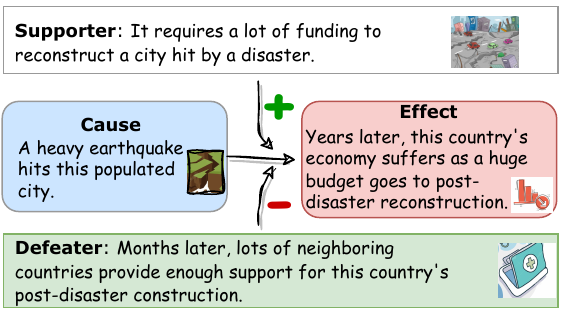}
\caption[]{A motivational example of defeasibility in causal reasoning. It consists of a cause-effect pair, a supporting argument~(supporter), and an opposing argument~(defeater) for the causal relationship. }
\label{fig:introduction:defeasibility}
\end{figure}

Despite the importance of causality, establishing a definite causal relationship between two events is inherently challenging as \textit{uncertainties} can influence the strength of the causality between the cause and the effect. 
As the time interval between the cause and the effect widens, these uncertainties tend to increase. 
These uncertainties include incomplete or unseen factors, changing situations, and contextual information. However, existing works on causal reasoning~\cite{qin2019counterfactual,feng-etal-2021-empowering,zhang2022rock,wang-etal-2023-cola} mainly focus on definite causality while ignoring the uncertainties inherent in causal relationships. 
\definecolor{positivegreen}{HTML}{009900} 
\definecolor{negativered}{HTML}{cc0100}  %

Motivated by this blank, we introduce the concept of defeasibility in causal reasoning. Formally, defeasibility in causal reasoning refers to situations wherein the causal relationship between the cause and the effect is justified, but supplementary information might strengthen or weaken these justifications. 
As the example shown in Figure~\ref{fig:introduction:defeasibility}, a supporter argument that ``It requires a lot of funding to reconstruct $\cdots$'' \ul{strengthens}~(\textcolor{positivegreen}{\textbf{+}}) the causal relationship between ``the earthquake'' and ``the country's economic decline''. 
On the other hand, a defeater argument that ``Months later, lots of neighboring countries provide enough support $\cdots$'' \ul{weakens}~(\textcolor{negativered}{\textbf{--}}) the causal relationship. 
With the capacity to understand defeasibility, humans can clearly perceive the change in the causal strength brought by supporters and defeaters.

However, prior benchmarks~\cite{roemmele2011choice,ning2018joint,sloman2005causal} on causal reasoning mostly focus on definite relationships, often overlooking the \textbf{defeasibility} of cause-effect pairs when uncertainties arise. To bridge this gap, we contribute the first benchmark for investigating defeasibility in causal reasoning: \shorttitle. As depicted in Figure~\ref{fig:introduction:defeasibility}, each sample in \shorttitle consists of a cause-effect pair, accompanied by its supporter argument~($A$) and defeater argument~($D$). $A$ and $D$ reinforce and undermine the causal relationship between the cause and the effect, respectively. We construct \shorttitle from examples across ten domains: environment, business, science, health, work, politics, education, sports, entertainment, and travel, and use our new benchmark to test how well existing large language models~(LLMs) can generate supporters and defeaters for causal pairs. Our experiments reveal that state-of-the-art pre-trained models, including GPT-3.5, lag behind humans by up to 4.5 and 10.7 points in generating correct supporters and defeaters, respectively, which emphasizes the significant challenges brought by \shorttitle.

Furthermore, due to the lack of appropriate benchmarks, it is difficult to determine whether existing metrics on qualifying causal strength can capture the change of causal strength brought by supporters and defeaters or not. An ideal metric should reflect the increase~(or decrease) in causal strength with the incorporation of supporters~(or defeaters). 
With the presence of supporters and defeaters, \shorttitle serves as an ideal touchstone for assessing the efficacy of existing metrics in capturing the causal strength change brought by supporters and defeaters. 
Our experiments demonstrate that existing cutting-edge metrics like ROCK~\cite{zhang2022rock} and CEQ~\cite{du2022care}, whose accuracy are both below 50\%, fail to accurately capture the causal strength change. 

To address this limitation, we propose a robust and versatile metric for measuring causal strength, known as CESAR, based on \underline{C}ausal \underline{E}mbedding a\underline{S}sociation with \underline{A}ttention \underline{R}ating~(CESAR).  CESAR builds upon a transformer-based model~\cite{devlin2019bert} with causal embeddings. The causal strength given by CESAR is calculated as a weighted aggregation of token-level causal strength, which is the association score between a token's causal embedding in the cause and its counterpart in the effect. The learned weighted coefficients guide CESAR to prioritize strong causal pairs like ``fire'' and ``burn''. 
From the experimental results, CESAR achieves a significant 69.7\% improvement in quantifying changes in causal strength resulting from supporters and defeaters. It also attains state-of-the-art performance with an 11.9\% improvement in distinguishing the correct hypotheses from incorrect ones, underscoring CESAR's versatility in various causal tasks.

In summary, we make four key contributions:
\begin{itemize}
    \item We contribute \shorttitle, the pioneering benchmark that emphasizes the often overlooked aspect of causal reasoning: defeasibility. It paves the road to systematically exploring defeasibility in causal reasoning. \shorttitle is available at \url{https://github.com/cui-shaobo/defeasibility-in-causality}. 
    \item With the presence of the supporters and defeaters, \shorttitle serves as a valuable yardstick for evaluating existing metrics on causal strength. We highlight the limitations of current causal strength metrics in capturing the changes in causal strength resulting from supporters and defeaters. 
    \item We propose CESAR, a robust and versatile metric for measuring causal strength. CESAR outperforms existing metrics like ROCK and CEQ, exhibiting a remarkable 69.7\% improvement in capturing the changes of causal strength brought by supporters and defeaters.
    \item Using \shorttitle, we assess the ability of existing LLMs to comprehend defeasibility in causal reasoning. The results show that even GPT-3.5 falls significantly short, lagging behind humans by 4.5 and 10.7 points in generating accurate supporters and defeaters, respectively. This underscores the significant challenges brought by \shorttitle. 
\end{itemize}
\section{Related Work}
\begin{table*}[htp!]
  \centering
  \resizebox{0.99\textwidth}{!}{
  \begin{tabular}{lp{1.5cm}lp{2cm}p{1.5cm}p{2cm}p{2cm}p{3.5cm}}
    \toprule
     & Annotation Unit & Size & Always Causality & \#Causality pairs  & Defeater & Supporter & Touchstone for causal strength metrics \\
    \midrule
    \rowcolor{mygray} \multicolumn{8}{c}{\textit{Commonsense causal reasoning datasets}}\\
    COPA \scriptsize{\cite{roemmele2011choice}} &  Sentence & 1,000 & \bfull & 1  & \bempty & \bempty & \bempty \\
    TCR\scriptsize{\cite{ning2018joint}} &  Sentence & 172 &\bfull & 1  & \bempty & \bempty & \bempty \\
    e-CARE\scriptsize{\cite{du2022care}} & Sentence & 21,324 & \bfull & 1  & \bempty & \bfull & \bempty \\
    \rowcolor{mygray} \multicolumn{8}{c}{\textit{Counterfactual commonsense reasoning datasets}} \\
    ART\scriptsize{\cite{bhagavatula2020abductive}} &  Sentence & 20,000 & \bempty & 1  & \bempty & \bempty & \bempty \\    TimeTravel\scriptsize{\cite{qin2019counterfactual}} &  Paragraph & 29,849 &\bempty & 2 & \bempty & \bempty & \bempty \\
    \rowcolor{mygray} \multicolumn{8}{c}{\textit{Defeasible inference datasets}}\\
    $\delta$-NLI \scriptsize{\cite{rudinger2020thinking}} &  Sentence & 9,986 & \bempty & N/A & \bfull & \bfull & \bempty \\
    \midrule
    \shorttitle & Sentence & 11,245 & \bfull & 2 & \bfull & \bfull & \bfull \\
    \bottomrule
  \end{tabular}
  }
  \caption{Comparison of \shorttitle and related datasets. \bfull means supported and \bempty means not.}
  \label{tab:related:comparison}
\end{table*}

\noindent\textbf{Comparison of \shorttitle with Related Datasets.\tbfspace} 
We present the comparison between \shorttitle and related datasets in Table~\ref{tab:related:comparison}. 
Most commonsense causal reasoning datasets like COPA~\cite{roemmele2011choice} and TCR~\cite{ning2018joint} focus on definite causality, ignoring the uncertainties inherent in the causal relationship. \shorttitle introduces defeasibility to cover uncertainties, enhancing its capacity to test models in defeasible causal reasoning. Counterfactual datasets like  TimeTravel~\cite{qin2019counterfactual} and ART~\cite{bhagavatula2020abductive} often lack consistently valid causal relationships, presenting causality pairs based on counterfactual events and lack the capacity to test the performance of existing causal strength metrics. 
In contrast, \shorttitle incorporates both supporters and defeaters and thus makes itself an idea touchstone for causal strength metrics.
Lastly, while \citet{rudinger2020thinking} define defeasible inference in natural language without always implying causality, \shorttitle emphasizes the causal relationship's defeasibility.

\noindent\textbf{Existing Evaluation Metrics on Causal Strength. \tbfspace} Previous literature~\cite{luo2016commonsense,du2022care,zhang2022rock} study the causal strength from different perspectives. \citet{du2022care} propose a metric named Causal Explanation Quality (CEQ) score based on word co-occurrence to measure if a given explanation could increase the causal strength between the cause and the effect. \citet{zhang2022rock} propose a theoretical framework named ROCK to measure the causal strength from the causal inference perspective. Details about existing causal strength metrics are present in Appendix~\ref{appendix:related_metrics}. 
\section{Task} \label{sec:task}
\noindent\textbf{Research Questions.\quad}
In this paper, we study two research questions to understand defeasibility: 
\begin{itemize}
    \item \textit{Research Question \RNum{1}}: How to estimate the strength of causality in the setting of defeasible causal reasoning? Specifically, can existing metrics on causal strength accurately capture the changes brought by supplementary information like supporters or defeaters in defeasible causal reasoning? 
    \item \textit{Research Question \RNum{2}}: Can language models generate correct defeasible arguments for given causal facts that can make the causality less justified or more justified? Specifically, we explore whether existing models can generate supporters or defeaters correctly. 
\end{itemize}
We answer Research Question \RNum{1} in \S \ref{sec:causal_strength} and Research Question \RNum{2} in \S \ref{sec:defeasibility_experiment}. 

\begin{figure}
  \begin{center}
  \includegraphics[width=0.28\textwidth]{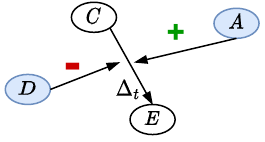}
  \end{center}
  \caption{The overview of the causal relationships in \shorttitle. 
  The symbol \textcolor{positivegreen}{$\textbf{+}$} indicates that the supporter $A$ strengthens the causal relationship between the cause $C$ and the effect $E$, while \textcolor{negativered}{\textbf{--}} signifies that the defeater $D$ weakens this relationship. The variable $\Delta_t$ denotes the time interval between the cause $C$ and the effect $E$. To ensure the practicality of identifying defeaters, this time interval is set to long time durations.
  }
  \label{fig:task:probabilistic}
\end{figure}

\vspace{0.2em}
\noindent\textbf{Estimating Causal Strength for Studying Research Question~\RNum{1}.\tbfspace} 
The causal strength between event $C$ and event $E$, denoted as $\causalstrength{C}{E}$, falls in $[0,1]$. It measures the intensity of the event $C$ causing/leading to the occurrence of event $E$. We present the overall causal relationship of defeasible causal reasoning in Figure~\ref{fig:task:probabilistic}. 

In the context of defeasible causal reasoning, an ideal metric on causal strength should meet the following requirements: (i) the estimated causal strength given by this metric should increase with the incorporation of supporters;  (ii) the estimated causal strength given by this metric should decrease with the incorporation of defeaters. 

\noindent\textbf{Supporter and Defeater Generation for Studying Research Question~\RNum{2}.\quad} 
Defeasibility is a fundamental concept in many fields.  In legal reasoning, defeasibility means a legal principle can be overridden by a competing principle. Defeasibility in causal reasoning implies that the validness of causality can be less or more justified by additional information like supporters and defeaters. With the definition of causal strength, the defeasibility in commonsense causal reasoning is represented as two constraints: 
\begin{equation}
\label{eq:strength}
\resizebox{0.85\hsize}{!}{
$
\begin{cases}
    \causalstrength{C}{E} - \causalstrength{(C \oplus A)}{E}  < 0 \\
    \causalstrength{C}{E} - \causalstrength{(C \oplus D)}{E}  > 0 \\
\end{cases}
$
}
\end{equation}
where $\oplus$ means the combination of two events.
$A$ and $D$ represent supporters and defeaters, respectively. 
The first constraint requires that the causal strength between the cause and the effect is strengthened by the supporter, while the second constraint requires that the defeater should weaken the causal strength. 

Given the cause and the effect, we ask the model to generate a supporter $A$ or a defeater $D$ that reinforces or diminishes the causal relationship between $C$ and $E$ as much as possible. Namely,
\begin{equation}
\resizebox{0.88\hsize}{!}{
$
\begin{cases}
    A = \argmax_{A} [ \causalstrength{(C \oplus A)}{E} - \causalstrength{C}{E}]\\
    D = \argmax_{D} [\causalstrength{C}{E} - \causalstrength{(C \oplus D)}{E}]\\
\end{cases}
$
}. 
\end{equation}

\section{\shorttitle} \label{sec:td-ccr}
\subsection{Overview of \shorttitle} \label{sec:dataset:overview}
Each instance in \shorttitle consists of four components: 
(1) a domain label from 10 domains including Environment, Business, Science/Technology, Health, Work, Politics, Education, Sports, Entertainment, and Travel;
(2) a cause-effect pair that is presented with a cause, its effect, and the time interval between the cause and the effect;
(3) a defeater argument that reduces the validness of or totally invalidates the causal relationship between the cause and the effect;
(4) a supporter argument that makes the causal relationship between the cause and the effect more justified. 
\begin{figure}[htp!]
\begin{center}
  \includegraphics[width=0.5\textwidth]{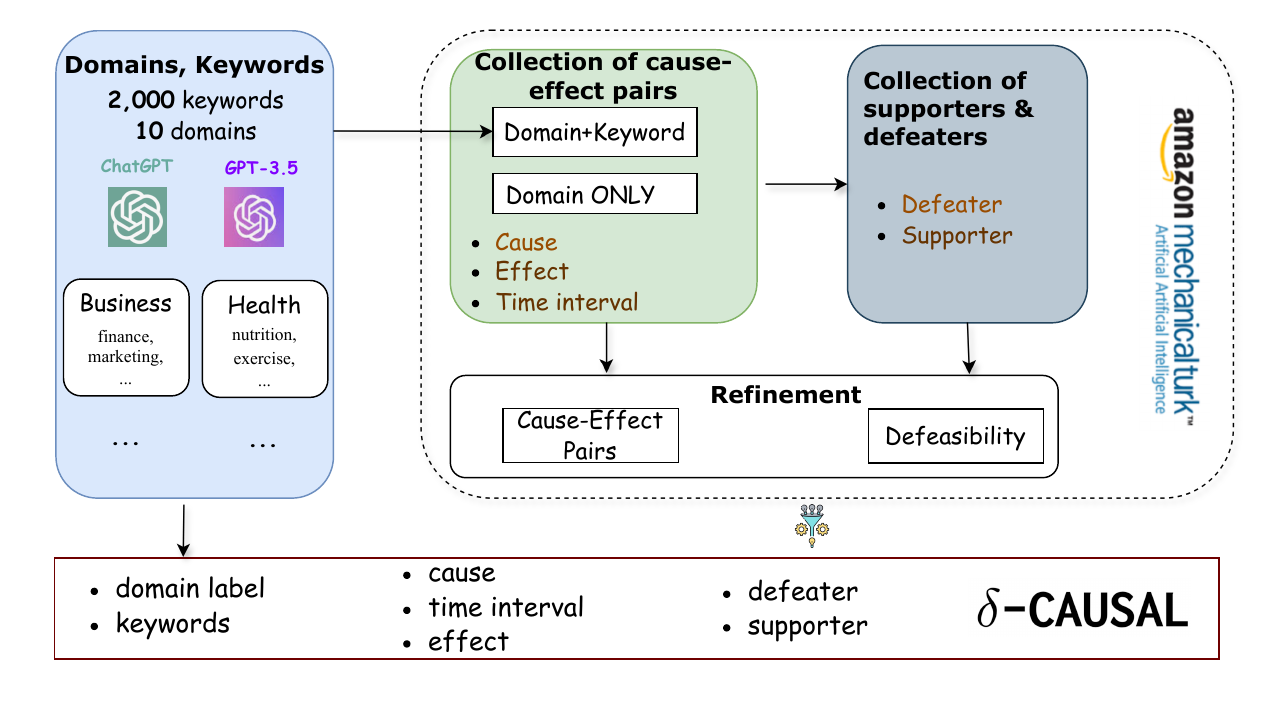}
  \end{center}
  \caption{Pipeline of the annotation and refinement procedures of \shorttitle.}
  \label{fig:dataset:pipeline}
\end{figure}

\subsection{Annotation of \shorttitle} \label{sec:dataset:annotation}
Figure~\ref{fig:dataset:pipeline} illustrates the data annotation and refinement pipeline. Initially, we gather keywords for each domain, which guide the annotation of cause-effect pairs on Amazon Mechanical Turk~(AMT). We also collect supporters and defeaters for these pairs on AMT. Each annotation step is paired with a refinement phase.

\begin{figure}[htp!]
    \centering
    \includegraphics[width=0.9\linewidth]{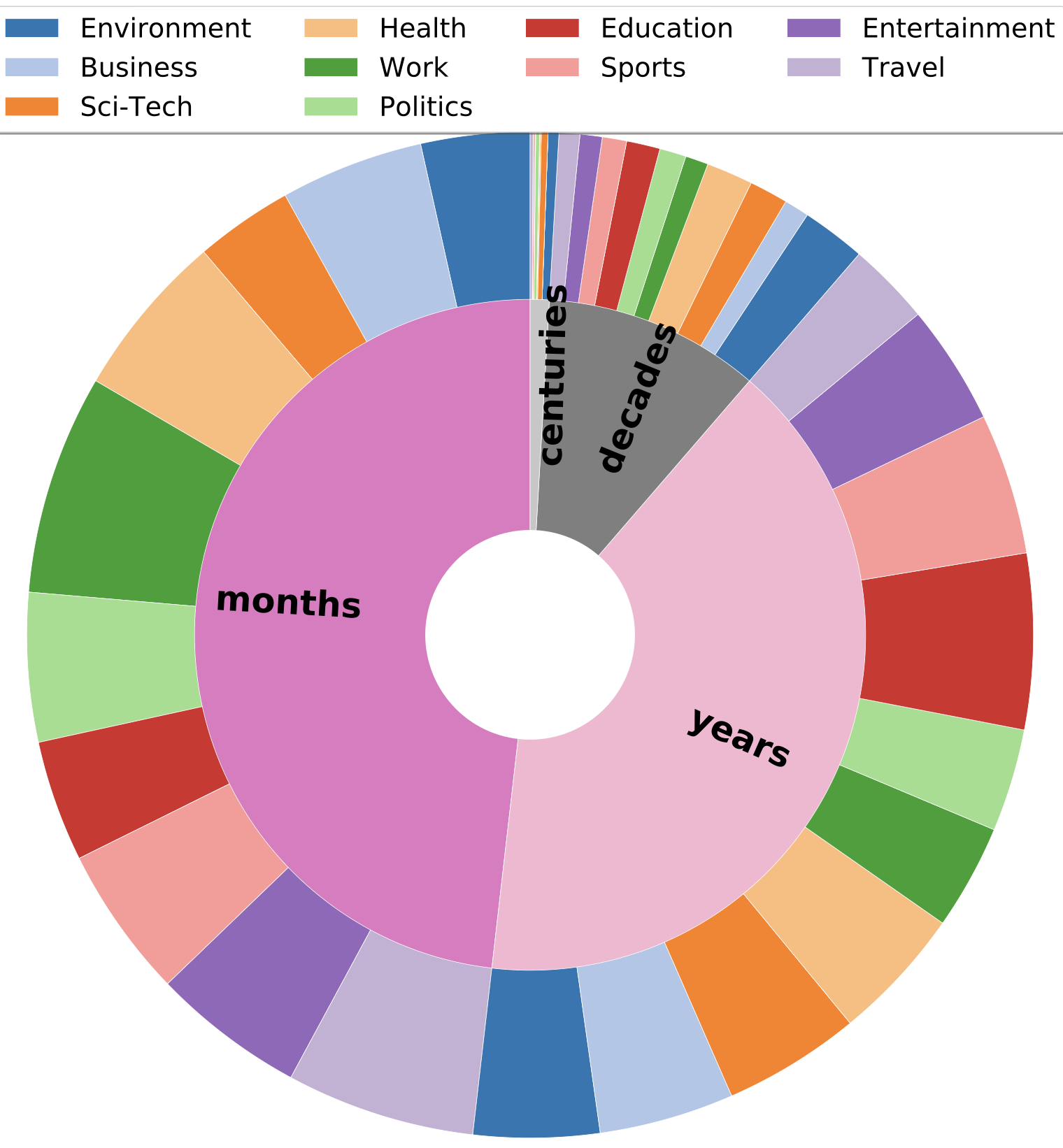}
    \caption{The distributions of time intervals and domains in \shorttitle. Different colors represent different time intervals~(inner circle) and domains~(outer circle). Detailed values and proportions are provided in Appendix~\ref{appendix:statistics:time_interval}. }
    \label{fig:dataset:distribution}
\end{figure}

\noindent\textbf{Phase \RNum{1}: Annotation of Cause-Effect Pairs.\tbfspace}
\textit{Without Keyword Hints.\tbfspace}
Annotators begin by selecting from ten domains, as detailed in Figure~\ref{fig:dataset:distribution}. Within the chosen domain, they first annotate a cause-effect pair. To ensure the feasibility of identifying defeaters, we impose restrictions on the selection of time intervals to relatively long-term periods. This practice is taken because short time intervals often make it challenging for annotators to identify and annotate potential defeating events effectively. Longer time intervals provide a broader temporal scope, allowing for the observation and annotation of more complex interactions and changes that might influence or negate the initial cause-effect relationship. This methodological choice enhances the richness and reliability of the annotated data by capturing a wider range of possible outcomes and influences over extended periods. Specifically, annotators must specify a time interval for the effect, with options including \texttt{months later}, \texttt{years later}, \texttt{decades later}, and \texttt{centuries later}. For more details on time labels, see Appendix~\ref{appendix:statistics:time_interval}.
\noindent\textit{With Keyword Hints.\tbfspace}
Our AMT data collection revealed limited topic variety within domains without keyword hints. For example, the Health domain often linked exercise to weight loss. To diversify our benchmark, we used GPT-3.5 to generate 200 keywords per domain (100 each from text-davinci-003 and ChatGPT) listed in Appendix~\ref{appendix:statistics:keywords}. Annotators then receive a keyword as a hint and craft a related cause-effect pair within the domain.

\noindent\textbf{Phase \RNum{2}: Annotation of Supporters and Defeaters.\tbfspace}
We ask annotators to write the supporter and the defeater simulataneously. The supporter can be a conceptual explanation or fact that supports the causal relationship, while the defeater should provide evidence for the opposite thesis of the effect or evidence that undercuts the effect~\cite{pollock1987defeasible}. A defeater can also be an event or a circumstance that makes the causality between the cause and the effect unjustified or less justified.

\subsection{Refinement of \shorttitle} \label{sec:dataset:refinement}
We enhance the benchmark quality through systematic refinement stages. Initially, we eliminate samples written randomly: several annotated examples either contain repetitive wording or merely echo the instructions. Subsequently, we undertake two Mechanical Turk (AMT) refinement phases: refining cause-effect pairs and refining supporters and defeaters. 

\noindent\textbf{Phase \RNum{3}: Cause-Effect Pair Refinement.}
We task annotators to assess the validity and timing of the cause-effect relationships. Each cause-effect pair is judged by three annotators. To ensure the quality, each assignment includes a gold cause-effect pair with a known label. Any assignments whose gold examples are incorrectly labeled are disregarded. We retain annotations if: (i) No annotation is discarded and a majority deem it true; (ii) one annotation is discarded for misjudging the gold example, while the remaining two validate the evaluated sample.

\noindent\textbf{Phase \RNum{4}: Refinement of Supporters and Defeaters.\tbfspace}
Three annotators determine whether the supporter/defeater enhances or diminishes the causal connection. Each task has a defeasibility pair with a known label and another for assessment. Similarly, as in Phase \RNum{3}, we only keep the assignments whose majority of the filtered votes are true. 

\subsection{Overall Quality of \shorttitle}\label{sec:dataset:quality}
In order to assess the quality of \shorttitle, we randomly select 200 samples from \shorttitle and ask three NLP experts~(see details in Appendix~\ref{appendix:experiments:experts}) to assess the validity of these samples from the following three perspectives:  validness of causality, supporter, and defeater. The assessment result is shown in Table~\ref{tab:quality}. We achieve an average accuracy $\text{Accuracy} \geq$ 92\% with a high agreement. This shows that \shorttitle is of good quality.

\subsection{Statistics of \shorttitle} \label{sec:dataset:statistics}
The statistics of \shorttitle are shown in Table~\ref{tab:statistics}. 
\begin{table}[htp!]
    \centering
    \resizebox{0.40\textwidth}{!}{
    \begin{tabular}{lr} 
       \toprule 
        & Statistics \\ 
       \midrule
       \rowcolor{mygray} \multicolumn{2}{c}{\textit{Overall}}\\
       \# Causality pairs & 8,080    \\
       \# Supporters & 11,245 \\
       \# Defeaters & 11,245 \\
       Train/Dev/Test & 7,000/2,276/1,969 \\
        \rowcolor{mygray} \multicolumn{2}{c}{\textit{Length of utterances}}\\
       Average length of causes & 9.50 \\
       Average length of effects & 9.60 \\
       Average length of supporters & 9.10 \\
       Average length of defeaters & 10.39 \\
      \bottomrule 
    \end{tabular}
    }
    \caption{Statistics of \shorttitle. More details about \shorttitle are shown in Appendix~\ref{appendix:statistics}.}
    \label{tab:statistics}
\end{table}
More details are in the Appendix. Specifically, Appendix~\ref{appendix:qualification} is the qualification rules for annotators. 
Appendix~\ref{appendix:collection} elaborates the gold examples and guidelines we use during dataset collection.
Appendix~\ref{appendix:statistics} is more about statistics of \shorttitle.

\section{Estimating Causal Strength} \label{sec:causal_strength}
\begin{figure*}[htp!]
\centering
\begin{subfigure}{.325\linewidth}
  \includegraphics[width=\linewidth]{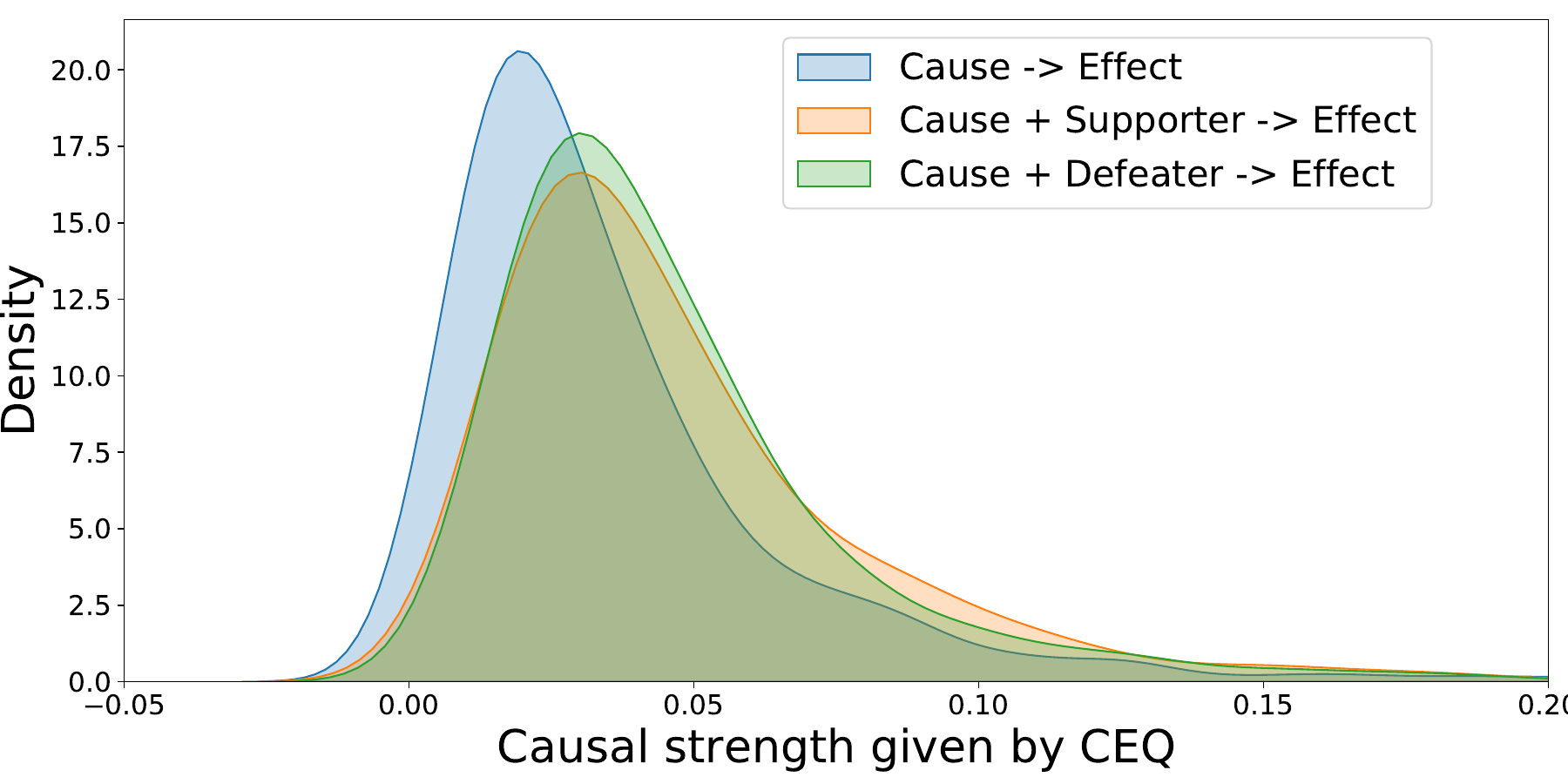}
  \caption{Shift of CEQ distribution. }
  \label{fig:appendix:shift_comparison:ceq}
\end{subfigure}
\begin{subfigure}{.325\linewidth}
  \includegraphics[width=\linewidth]{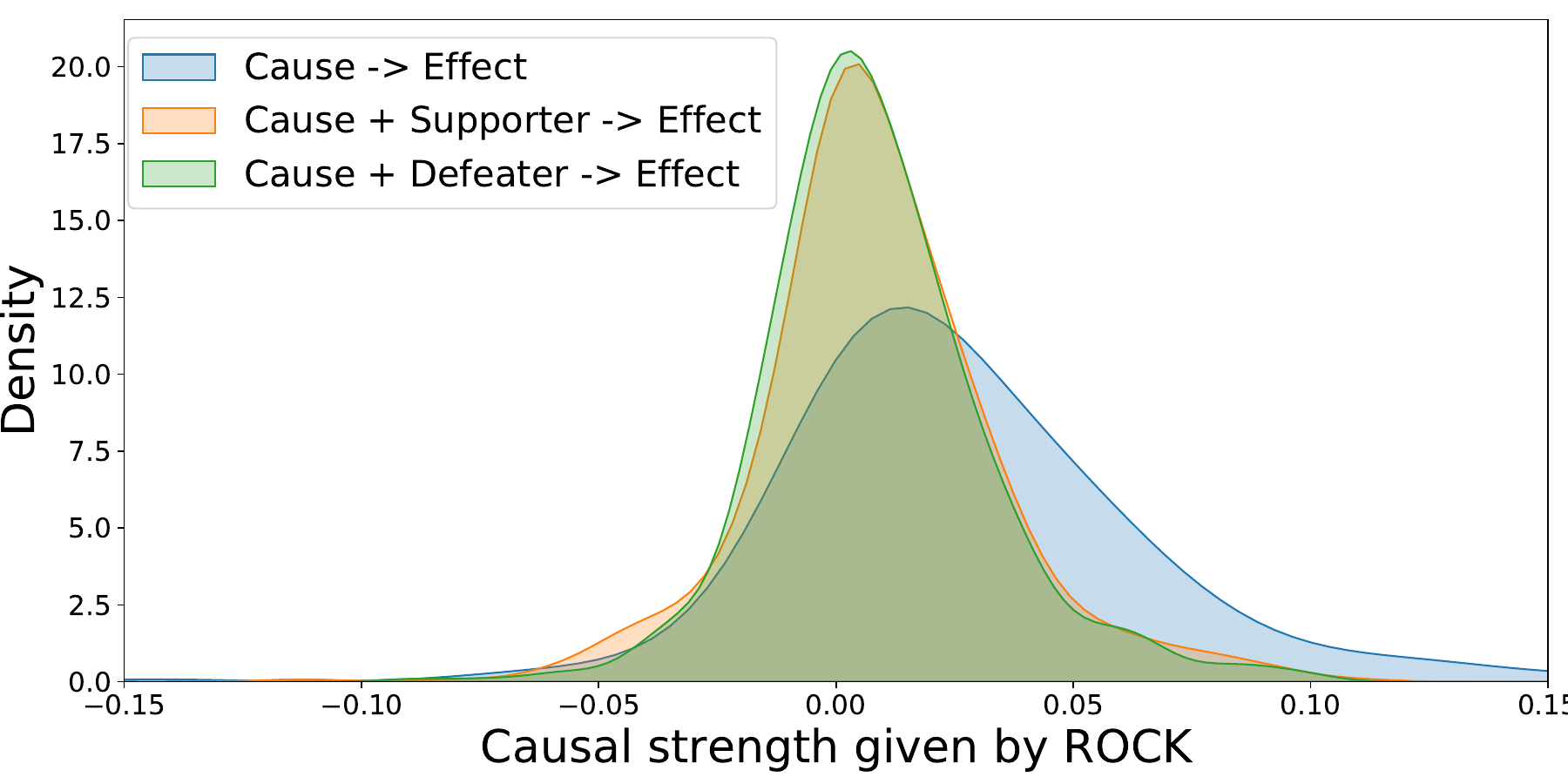}
  \caption{Shift of ROCK distribution.}
  \label{fig:appendix:shift_comparison:rock}
\end{subfigure}
\begin{subfigure}{.325\linewidth}
  \includegraphics[width=\linewidth]{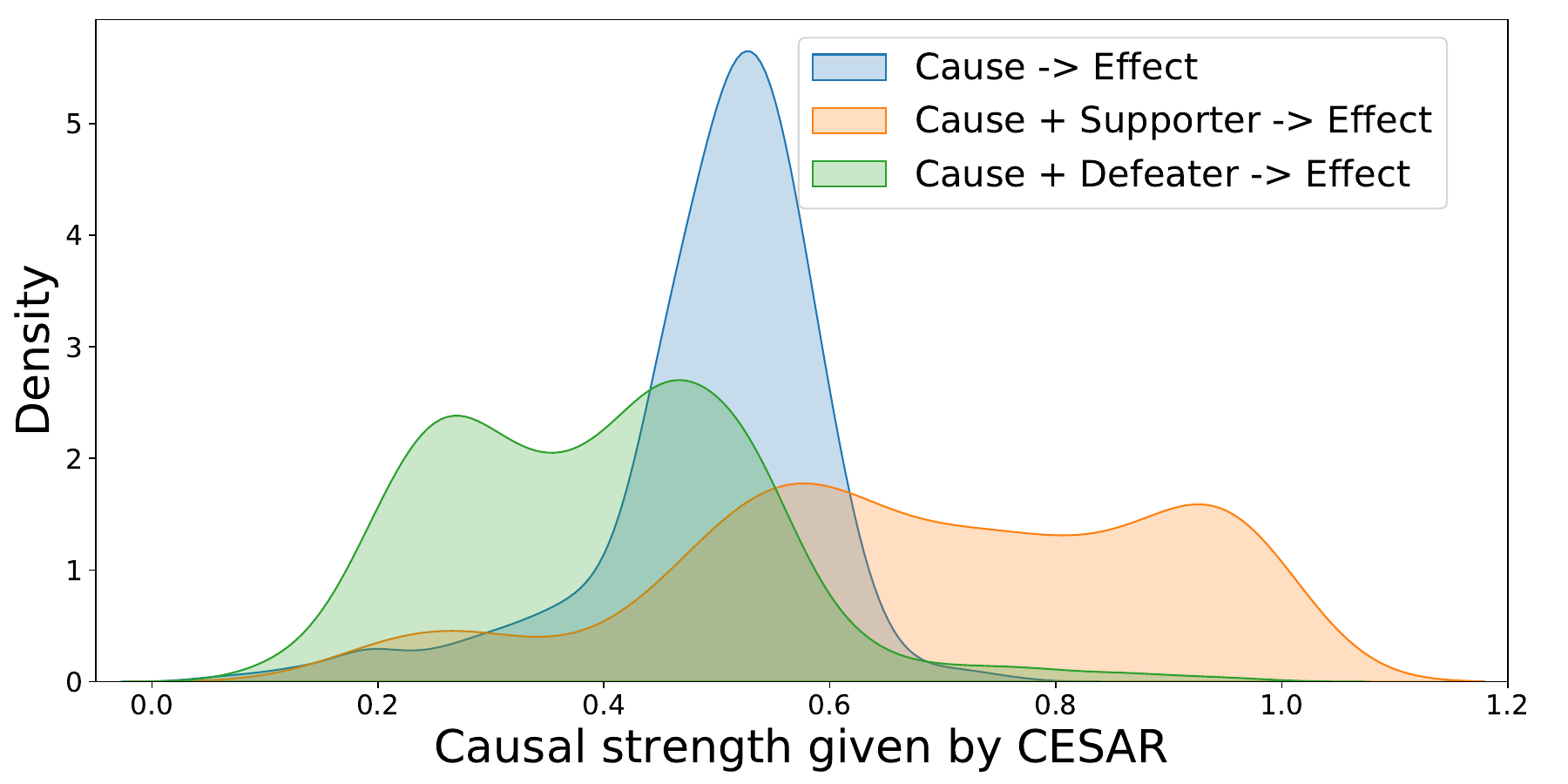}
  \caption{Shift of CESAR distribution.}
  \label{fig:appendix:shift_comparison:cesar}
\end{subfigure}
\caption{
The shifts in causal strength distributions facilitated by CEQ~(left), ROCK~(middle), and CESAR~(right) with the incorporation of supporters and defeaters are illustrated in \shorttitle. These curves utilize kernel density estimation~\cite{parzen1962estimation} to depict the data distribution as a continuous probability density curve. Notably, only CESAR effectively captures the variations in causal strength triggered by the inclusion of supporters and defeaters; specifically, the causal strength distribution shifts to the right with supporters and to the left with defeaters.
}
\label{fig:appendix:shift_comparison}
\end{figure*}
In this section, we address Research Question \RNum{1} on causal strength. We first highlight the limitations of current metrics in \S~\ref{sec:discussions:challenges}. Then, we detail the definition, comparison with other metrics, versatility, and case study of CESAR from \S~\ref{sec:discussions:cesar} to \S~\ref{sec:discussions:case}. 
\subsection{Limitations of Existing Metrics} \label{sec:discussions:challenges}
\shorttitle works as a solid touchstone to test existing metrics for evaluating causal strength. As mentioned earlier, in \shorttitle, a supporter is expected to increase the causal strength between the cause and the effect, while a defeater is expected to decrease the causal strength, as depicted in Equation~\eqref{eq:strength}. 

\begin{table}[htp!]
    \centering
    \resizebox{\linewidth}{!}{
    \begin{tabular}{p{2.5cm}p{1.5cm}p{1.5cm}p{2.0cm}}
      \toprule 
       & Supporter & Defeater & Geometric mean\\
      \midrule
      CEQ & 83.1 &  17.5 & 38.1\\
      ROCK &  32.5 & 68.6 & 47.2 \\
      \midrule
      CESAR~(ours) & \textbf{84.6} & \textbf{75.8} &  \textbf{80.1} \\
      \bottomrule
    \end{tabular}
  }
  \caption{Accuracy of causal strength metrics on \shorttitle: For supporters, correct predictions occur when the metric assigns a higher score to the cause-supporter combination. For defeaters, predictions are correct if the metric assigns a lower score to the cause-defeater combination. The geometric mean is calculated based on the accuracy of supporters and defeaters. 
  }
  \label{tab:causal_metrics}
\end{table}

The accuracy of metrics on \shorttitle for capturing causal strength changes by supporters and defeaters is detailed in Table~\ref{tab:causal_metrics}.
CEQ sees 83.1\% of supporters as strengtheners, but wrongly views 82.5\% of defeaters as such. Conversely, ROCK incorrectly sees 67.5\% of supporters as weakeners.
These results highlight the limitations of existing metrics and emphasize the need to develop more robust evaluation metrics for causal strength.

\begin{figure*}[!t]
    \centering
    \includegraphics[width=0.99\textwidth]{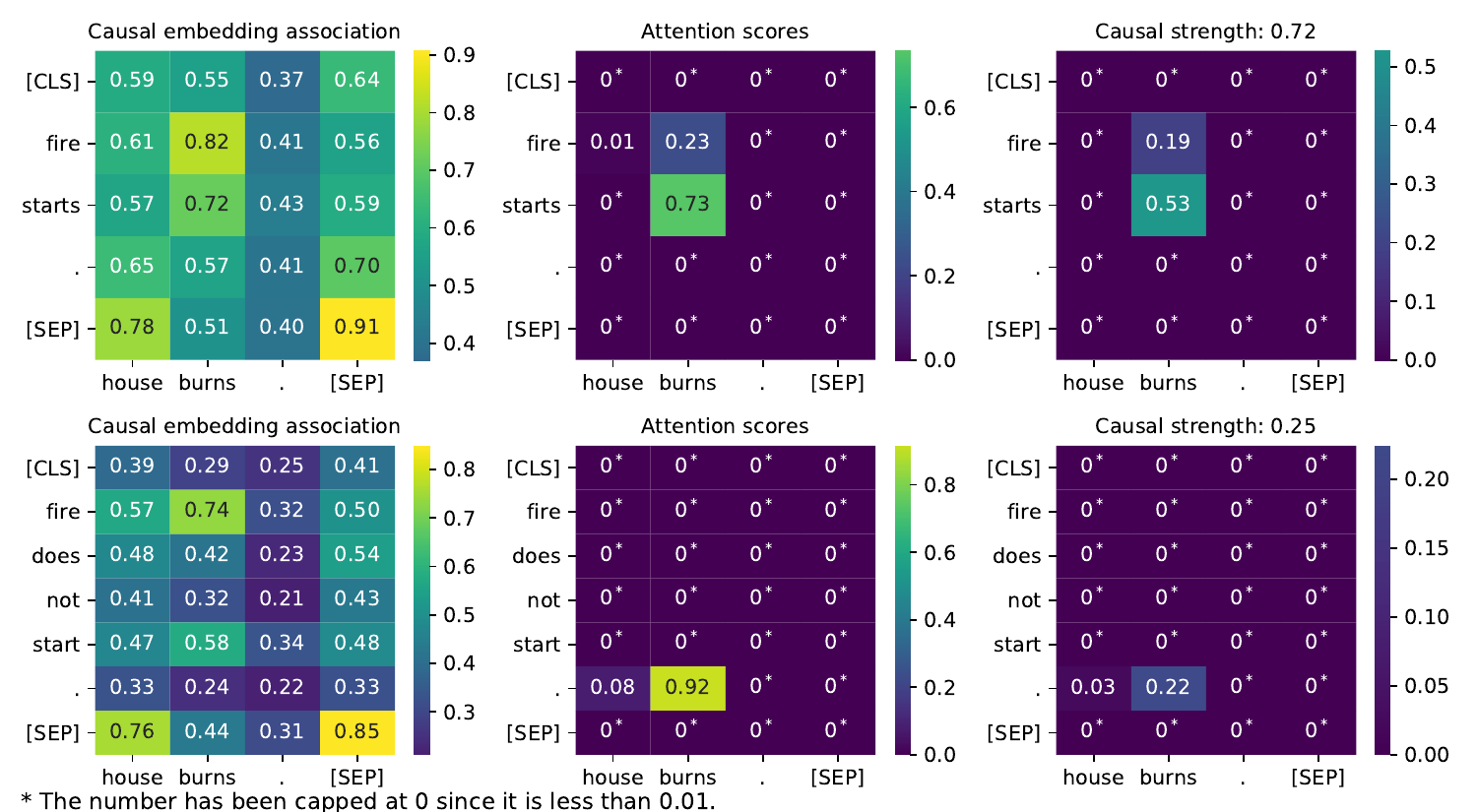}
    \caption{ We define three matrices based on Equation~\eqref{eq:cesar}. 
    (i) Causal Embedding Association Matrix~($\mathbf{M}$) in the left column: $M_{ij}$ measures the causal association between embeddings $c_i$ and $e_j$ as $M_{ij} = \frac{|c_i^T e_j|}{\Vert c_i\rVert\Vert e_j\rVert}$. 
    (ii) Causality-Aware Attention Matrix~($\mathbf{A}$) in the middle column: $\mathbf{A}$ stores the attention weights of each token pair given by $A_{ij} = a_{ij}$. 
    (iii) Causal Strength Matrix~($\mathbf{S}$) in the right column: $\mathbf{S}$ is the Hadamard product between $\mathbf{M}$ and $\mathbf{A}$. That is, $S_{ij} = M_{ij} A_{ij} = a_{ij} \frac{|c_i^T e_j|}{\Vert c_i\rVert\Vert e_j\rVert}$.
    The upper row of this figure shows the values of $\mathbf{M}$, $\mathbf{A}$, and $\mathbf{S}$ derived from the cause-effect pair wherein the cause is ``Fire starts.'' and the effect is ``House burns.'' 
    Conversely, the lower row presents the values of $\mathbf{M}$, $\mathbf{A}$, and $\mathbf{S}$ when the cause is ``Fire does not start.'' and the effect is ``House burns.''. The y-axis denotes cause tokens, and the x-axis represents effect tokens.
    The causal strength, calculated from Equation~\eqref{eq:cesar}, is shown in the title of the rightmost matrices for both pairs. 
    }
    \label{fig:appendix:cause_matrix}
\end{figure*}
\subsection{CESAR: \texorpdfstring{\underline{C}ausal \underline{E}mbedding A\underline{S}sociation with \underline{A}ttention \underline{R}ating}{Causal Embedding Similarity with Attention Rating} on Causal Strength}
\label{sec:discussions:cesar}
\noindent\textbf{Motivation: Causality-aware Embedding and Attention.\tbfspace}
The causal strength between two events can be quantified as the weighted aggregation of the causal relationship between tokens within these events~\cite{luo2016commonsense}. For instance, in events ``Fire starts'' and ``House burns'', ``fire'' and ``burns'' drive a causal relationship. 
Inspired by BERTScore \cite{zhang2019bertscore}, 
we fine-tune BERT embeddings to capture token-level causality so that tokens delivering a strong causal relationship, like ``fire'' and ``burns'', have embeddings that are highly associated. 
The motivation for the attention mechanism is that we wish to place less attention on pairs that consist of causality irrelevant words(e.g., stop words)  and more attention on pairs involved in a strong causal relationship like ``fire'' and ``burns''.

\noindent\textbf{Attention Computation.\tbfspace} On top of the BERT model, we add a customized attention layer that identifies the important token pairs for evaluating the causal strength. The attention scores for token pairs are calculated through a specifically adjusted cross-attention layer on the top of the BERT model. 
Let $d$ be the dimension of BERT model; $C$ and $E$ be tokenized to $n$ and $m$ tokens respectively, i.e., $\mathbf{C} \in \mathbb{R}^{n \times d}$ and $\mathbf{E} \in \mathbb{R}^{m \times d}$.
We compute query vectors as $\mathbf{Q} = \mathbf{C}\mathbf{W}_q$ and key vectors as $\mathbf{K} = \mathbf{E}\mathbf{W}_k$, where $\mathbf{W}_q, \mathbf{W}_k \in \mathbb{R}^{d \times d}$. 
Then, the matrix of attention scores for token pairs is calculated as $\mathbf{A} = \text{softmax}\left(\mathbf{QK}^T\right)$, where $\mathbf{A} \in \mathbb{R}^{n \times m}$, and softmax is performed over all values of the matrix, i.e., $\text{softmax}({A}_{ij}) = \frac{\exp{({A}_{ij})}}{\sum_{i,j} \exp{(A_{ij}})}$.

\noindent\textbf{Weighted Average of Causal Embedding Association. \tbfspace} We propose the following formula for computing the causal strength between $C$ and $E$:
\begin{equation}
\label{eq:cesar}
      \causalstrength{C}{E} = \sum_{i=1}^{n} \sum_{j=1}^{m} a_{ij}\frac{|c_i^T e_j|}{\Vert c_i\rVert\Vert e_j\rVert}
\end{equation}
where $c_1, \, c_2, \dots, \, c_n$ and $e_1, \, e_2, \dots, \, e_m$ are causal embeddings of tokens in $C$ and $E$, respectively. These embeddings are generated by the last hidden layer of fine-tuned BERT. The weight $a_{ij}$ is the attention put on each token pair of $c_i$ and $e_j$ such that $\sum_{i, j} a_{ij}=1$. We compute the absolute cosine similarity between causal embeddings for each pair of tokens and calculate the score as the average of these token-level causal associations weighted by the learned attention.

\noindent\textbf{Training Procedure. \tbfspace} We train the CESAR metric on the augmented e-CARE dataset~\cite{du2022care}. It contains cause-effect pairs with a conceptual explanation designed to increase the pair's causal strength. Specifically, we set $\causalstrength{C}{E}$ to 0.7 and $\causalstrength{C \oplus H}{E}$ to 1.0 where $H$ is the explanation for the causal relationship between $C$ and $E$. For pairs with no causal relationship, we set the causal strength to 0.0. Further, we use ChatGPT to generate opposites of the conceptual explanations provided in e-CARE and set $\causalstrength{C \oplus \neg H}{E}$ to 0.2 where $\neg H$ is an opposite of the conceptual explanation for $C$ and $E$. We train CESAR for $4$ epochs with AdamW optimizer~\cite{loshchilov2017decoupled} with a linear scheduler, learning rate 1e-5, and MSE loss. 

\noindent\textbf{Experimental Results. \tbfspace}
The experimental results are shown in Table~\ref{tab:causal_metrics}. It is evident that CESAR outperforms CEQ and ROCK significantly on both supporters and defeaters, achieving an accuracy of 84.6\% and 75.8\%, respectively. 
Furthermore, we use the geometric mean of the accuracy achieved in supporters and defeaters as the index of the overall performance for each metric. From Table~\ref{tab:causal_metrics}, we can see that both CEQ and ROCK attain a geometric mean accuracy of less than 50\%, In contrast to this poor performance, our proposed CESAR obtains a superior accuracy of 80.1\%. 

\subsection{Shift of Causal Strength Distributions with Supporters and Defeaters} \label{sec:discussions:comparison}
In Figure~\ref{fig:appendix:shift_comparison}, we plot the shift of causal strength distribution with the incorporation of supporters and defeaters for CEQ, ROCK, and CESAR~(ours).
For CEQ in Figure~\ref{fig:appendix:shift_comparison:ceq},
both supporter and defeater distributions shift rightward.
This means CEQ perceives any supplementary information as supporting the original cause-effect relationship, regardless of its actual impact on causal strength.
For ROCK in Figure~\ref{fig:appendix:shift_comparison:rock}, both distributions lean left, suggesting ROCK sees all supplementary information as opposing the original cause-effect relationship. 

For CESAR, the supporter distribution shifts right while the defeater distribution shifts left. 
This is the anticipated behavior for a good causal strength metric, capturing the contrasting effects of supporters and defeaters.

\subsection{Versatility of CESAR} \label{sec:discussions:generalness}
To additionally investigate the versatility of CESAR, we test CESAR on COPA~\cite{roemmele2011choice} and show the results in Table~\ref{tab:generalness_cesar_copa}. The accuracy reflects whether the tested causal strength metric gives a larger value to the causal strength of true cause-effect pairs than that of the false cause-effect pairs. 
CESAR achieves an accuracy of 70.7\%, once again significantly outperforming both ROCK and CEQ. This proves the generalness of CESAR in estimating the causal strength. 
\begin{table}[htp!]
    \centering
    \resizebox{0.43\textwidth}{!}{
    \begin{tabular}{l|lll}
      \toprule 
       Metrics & CEQ & ROCK & CESAR~(ours)\\
      \midrule
      Accuracy & 57.8 & 63.2 & \textbf{70.7} \\
      \bottomrule
    \end{tabular}
  }
  \caption{Results of CESAR's versatility, which is about distinguishing the true cause-effect pairs from the false cause-effect pairs on COPA~\cite{roemmele2011choice}.}
  \label{tab:generalness_cesar_copa}
\end{table}
\subsection{Case Study of CESAR} \label{sec:discussions:case}
Figure~\ref{fig:appendix:cause_matrix} illustrates the adaptation of values in Equation~\eqref{eq:cesar} when the cause is altered. 
In the upper row with inputs $C = \text{``Fire starts.''}$ and $E = \text{``House burns.''}$, the token causal embeddings of ``fire'' and ``burns'' have a high association score of 0.82. Also ``starts'' and ``burns'' demonstrate a strong association score of 0.72. Interestingly, a notable amount of attention is paid to the latter pair, whose association score is a key determinant of the causal strength score. Using Equation~\eqref{eq:cesar}, we obtain a causal strength of 0.72, which signifies a strong causal relationship between $C$ and $E$.
In the lower row, with a modified cause $C = \text{``Fire does not start.''}$ and the same effect $E$, 
the token pair associations of ``fire'' and ``burns'', and ``start'' and ``burns'' remain high. 
However, CESAR's attention mechanism adjusts the importance of tokens in terms of the causal relationship between two sentences in the right direction.
Namely, the causal strength undergoes a reduction of over 65\%, resulting in a score of 0.25 that indicates a weak causal relationship.

Details on setup, score computation, preparation of the training data, and an extensive ablation study of CESAR are in Appendix~\ref{appendix:cesar}.

\section{Supporter and Defeater Generation} \label{sec:defeasibility_experiment}
In this section, we answer Research Question \RNum{2} about existing SOTA models' ability in supporter and defeater generation by extensive experiments. 
\begin{table}[htp!]
    \resizebox{0.46\textwidth}{!}{
    \begin{tabular}{lllllp{2.0cm}} 
      \toprule 
      Model & BLEU & 
      METEOR & 
      ROUGE-L &
      CIDEr & 
      BERT-Score  \\
      \midrule
      \rowcolor{mygray} \multicolumn{6}{c}{\textit{Supporter generation}}\\
      BART & 7.71  & 12.90 & 16.72 & 0.397 & 54.0 \\
      T5 &  6.92  & 11.89 & 15.94 & 0.360 & 52.5 \\
      T5-large &  7.90  & 12.55 & 17.27 & 0.440 & 54.2 \\
      GPT-2 &  6.62 & 11.81 & 14.95 & 0.357 & 52.4 \\   
      GPT-3.5 & 3.17 & 10.97 & \hspace{0.48em}9.93 & 0.094 & 48.0 \\
      \rowcolor{mygray} \multicolumn{6}{c}{\textit{Defeater generation}}\\
      BART & 7.53 & 11.15 & 16.63 & 0.345 & 51.8 \\
      T5 & 6.83 & 10.89 & 15.83 & 0.279 & 51.5 \\
      T5-large & 7.37 & 10.90 & 16.48 & 0.325 & 52.1 \\
      GPT-2 & 6.71 & 10.38 & 15.32 & 0.257 & 50.9 \\
      GPT-3.5 & 5.24 & 10.86 & 15.27 & 0.205 & 50.0 \\
      \bottomrule 
    \end{tabular}
    }
    \caption{Results of supporter and defeater generation with different language models.}
    \label{tab:results}
\end{table}

\noindent\textbf{Setup.\tbfspace} We finetune generative pre-trained language models BART~\cite{lewis2020bart}, T5~\cite{raffel2020exploring}, T5-large, and GPT-2~\cite{radford2019language}. These models take the concatenation of the cause and the effect as the input. The output is the supporter or the defeater. 
See details about the baselines and experimental setup in Appendix~\ref{appendix:experiments}. We automatically evaluate the generated supporters/defeaters using BLEU~($n=2$)~\cite{papineni2002bleu}, METEOR~\cite{banerjee2005meteor}, ROUGE-L~\cite{lin2004rouge}, CIDEr~\cite{vedantam2015cider}, and BERT-Score~\cite{zhang2019bertscore}. 
\noindent\textbf{Results and Analysis.\tbfspace} The results of the supporter and defeater generation are shown in Table~\ref{tab:results}. 
It shows that existing LLMs perform unsatisfyingly in generating supporters and defeaters. Notably, we see that even GPT-3.5~(one-shot performance) does not achieve a high score based on existing generation metrics. This implies that existing metrics cannot adequately address the evaluation objective. Motivated by this, we employ a human evaluation to assess the quality of the generated supporters and defeaters from humans and GPT-3.5.
\begin{table}[htp!]
    \centering
    \resizebox{0.4\textwidth}{!}{ 
    \begin{tabular}{p{3cm}p{2cm}p{2cm}} 
       \toprule 
        & \% validness & \% agreement \\ 
       \midrule
        \rowcolor{mygray} \multicolumn{3}{c}{\textit{Quality of cause-effect pairs}}\\
        & 94.67 & 89.00  \\
        \rowcolor{mygray} \multicolumn{3}{c}{\textit{Quality of supporters}}\\
       Human & 92.67 & 83.00 \\
       GPT-3.5 & 88.17 & 73.50  \\ 
       \rowcolor{mygray} \multicolumn{3}{c}{\textit{Quality of defeaters}}\\
       Human & 94.50 & 81.50  \\
       GPT-3.5 & 83.83 & 69.50 \\
      \bottomrule 
    \end{tabular}
    }
    \caption{Comparison between humans and GPT-3.5 on supporter and defeater generation by human evaluation. }
    \label{tab:quality}
\end{table}

\noindent\textbf{Comparison between Humans and GPT-3.5 on Defeater and Supporter Generation by Human Evaluation.\tbfspace} 
Three NLP experts~(details in Appendix~\ref{appendix:experiments:experts}) are asked to judge the validness of the generated supporters and defeaters from GPT-3.5 and annotators. 
From the results in Table~\ref{tab:quality}, we can observe that existing large pre-trained models can not generate supporters and defeaters well. Even the large language model GPT-3.5 still lags behind humans by 4.5 points, i.e., 88.17\%~(GPT-3.5) vs. 92.67\%~(humans), in generating credible supporters. Even worse, it lags 10.7 points behind humans, i.e., 83.83\%~(GPT-3.5) vs. 94.50\%~(humans) in generating plausible defeaters. Sometimes, the model negates the effect rather than providing supplementary information to make it less justified. This demonstrates the challenges posed by \shorttitle. 
\section{Conclusions}
Our paper introduces \shorttitle, a pioneering benchmark that focuses on the often overlooked aspect of causal reasoning: defeasibility. 
Even state-of-the-art models like GPT-3.5 fall significantly short compared to human performance in understanding defeasibility shown by \shorttitle. 
We further demonstrate the limitations of current causal strength metrics in capturing causal strength changes brought by supporters and defeaters. To circumvent these limitations, we propose CESAR, a robust metric that outperforms existing measures by a remarkable 69.7\% improvement in capturing these changes.
Our research contributes to the advancement of causal reasoning by emphasizing defeasibility and providing valuable insights for improving language models' understanding of nuanced causal relationships. 
This work establishes a foundation for future studies on developing more advanced defeasible causal reasoning systems.

\section*{Acknowledgements}
We are grateful to Ping, Wanhao, and Mathieu for their invaluable assistance and inspiring discussions related to the writing and experimental design of this paper. We also gratefully acknowledge the financial and IT support provided by EPFL. 
AB gratefully acknowledges the support of the Swiss National Science Foundation (No. 215390), Innosuisse (PFFS-21-29), the EPFL Science Seed Fund, the EPFL Center for Imaging, Sony Group Corporation, and the Allen Institute for AI.

\section*{Limitations}
Despite our work's significant contributions, such as providing the first benchmark dataset for defeasible causal reasoning and introducing a novel causal strength metric, several acknowledged limitations still need to be addressed.
First and foremost, the causal strength change resulting from supporters and defeaters is currently described qualitatively rather than quantitatively. This limitation hinders the quantitative application of \shorttitle as it becomes challenging to precisely assess the exact magnitude of the causal strength change caused by supporters and defeaters. Consequently, it is difficult to use \shorttitle to quantify and measure the precise impact of these factors on causal reasoning.
Secondly, the domains covered by \shorttitle remain limited in scope. Expanding the applicability of \shorttitle to include other domains such as medicine or chemistry would enhance its versatility and make it more relevant to a broader range of research and practical applications. By incorporating additional domains in the future, we can evaluate the performance and effectiveness of \shorttitle in various contexts, ensuring its robustness and generalizability.
In conclusion, while our work has made notable contributions, it is essential to address these known limitations to enhance the quantitative usage and domain coverage. By doing so, we can advance the field of defeasible causal reasoning and strengthen the practical utility of our proposed metric on causal strength. 

\section*{Ethical Considerations}
We foresee no major ethical concerns for this work. 
As we know, causality contains various aspects of daily life. Bad things lead to negative results. But we take the following steps to make sure that the \shorttitle contains harmful/toxic content as little as possible. Firstly, a clear and understandable guide is given for annotations. After that, all of these annotated examples are followed up with a refinement process to filter out bad examples. Finally, we manually check whether the annotations contain keywords that convey harmful, toxic, violent, or erotic meanings. However, we acknowledge that these steps are not perfect.   
\shorttitle is under MIT License. Our paper involves other datasets, including e-CARE~\cite{du2022care}, which is under MIT License, and COPA~\cite{roemmele2011choice}, which is under BSD 2-Clause License. 


\bibliography{custom}
\clearpage
\appendix

\section{Qualification of Annotators} 
\label{appendix:qualification}
For the collection of \shorttitle, we expect it to have diversity while maintaining good quality. To achieve this requirement, we need to include more annotators while ensuring the quality of their annotations. 

In the collection process, we set the qualification that the annotators should have a HIT acceptance rate greater than 97 and a number of HITs approved greater than 10,000, which is a more general qualification rule to incorporate more annotators into our dataset collection process. This step ensures us a diversified dataset. During the refinement step, we maintain a group of annotators~(38 annotators) who understand our task well, we got this qualified annotator list by doing a toy collection of over 500 samples. Additionally,  to ensure that we can justify the quality of each refinement assignment, we assign each assignment to three annotators. Each assignment is composed of a golden example, which we know is true or false, and the example needs testing. Besides, as we focus on the English corpus, we set the country of residence to the USA and UK. 

In summary, we use general qualifications in the collection and more specific qualified groups in the refinement. In this way, we emphasize diversity in the collection and accuracy in the refinement.

\section{Details of Dataset Collection} \label{appendix:collection}
In this section, we present the guidelines and illustrative examples we give the annotators in detail. 
\subsection{Annotation of Cause-Effect Pairs}
\noindent\textbf{Guideline of Annotation.\tbfspace} Here we take the annotation of cause-effect pairs with hints of keywords as the instance. The annotation without keywords is similar except that the keywords are not given. 
For each annotating example~(HIT in Amazon Turk), we assign a domain and one keyword to the annotator and ask them to write a cause and its effect. To ensure that they write the  effect, we explicitly require them to select a time interval for the effect from \texttt{months later, years later, decades later} and \texttt{centuries later}. 

\noindent\textbf{Illustrative Examples.\tbfspace}
The illustrative examples we give for each domain for the cause-effect pairs are as follows~\footnote{Only the cases with hints of keywords are given. The cases without hints of keywords are similar}:

\noindent\texttt{Domain of Environment} 
\begin{enumerate}
    \item \textit{keyword}: natural disaster. \\
        \textit{Cause}: A tsunami hits the west coast. \\
        \textit{Effect}: Years later, homelessness and mental health issues arise.
    \item \textit{keyword}: natural disaster. \\
        \textit{Cause}: An earthquake happens in the city. \\
        \textit{Effect}: Years later, the city commemorates earthquake victims with charity events. 
\end{enumerate}

\noindent\texttt{Domain of Politics} 
\begin{enumerate}
    \item \textit{keyword}: political-party. \\
        \textit{Cause}: Tom founded this party with the hope of leading people to a better life. \\
        \textit{Effect}: Centuries later, it becomes the world's oldest active political party.
    \item \textit{keyword}: election. \\
        \textit{Cause}: The senator made a racist remark. \\
        \textit{Effect}: Months later, the senator's remarks cost him an election. 
\end{enumerate}

\noindent\texttt{Domain of Travel} 
\begin{enumerate}
    \item \textit{keyword}: resort. \\
        \textit{Cause}: Tourists throw rubbish everywhere at the scenic spot. \\
        \textit{Effect}: Years later, fewer and fewer tourists go to this scenic spot.
    \item \textit{keyword}: trip. \\
        \textit{Cause}: A young tourist is very happy after visiting Lausanne. \\
        \textit{Effect}: Decades later, this tourist revisits the old place, remembering the good old days. 
\end{enumerate}

\noindent\texttt{Domain of Entertainment} 
\begin{enumerate}
    \item \textit{keyword}: art. \\
        \textit{Cause}: The artist signs a recording contract. \\
        \textit{Effect}: Years later, the artist becomes a star and is popular among people.
    \item \textit{keyword}: movies. \\
        \textit{Cause}: The plot of the newly released film is very intriguing. \\
        \textit{Effect}: Decades later, this movie has been remade several times, and is known around the world. 
\end{enumerate}

\noindent\texttt{Domain of Sports} 
\begin{enumerate}
    \item \textit{keyword}: soccer. \\
        \textit{Cause}: The government is determined to make deep changes to professionalize its soccer and bring players closer to the global standard. \\
        \textit{Effect}: Decades later, the soccer team won the World Cup finally.
    \item \textit{keyword}: doping. \\
        \textit{Cause}: The athlete has been caught doping in the Olympics. \\
        \textit{Effect}: Years later, the athlete falls sick and retires from sports.
\end{enumerate}

\noindent\texttt{Domain of Education} 
\begin{enumerate}
    \item \textit{keyword}: school. \\
        \textit{Cause}: Tom's parents decide to let Tom enroll in a famous but expensive school. \\
        \textit{Effect}: Years later, Tom is well-educated and is thankful for his parents' efforts.
    \item \textit{keyword}: major. \\
        \textit{Cause}: Tom changes his major from mathematics to computer science. \\
        \textit{Effect}: Decades later, Tom becomes a senior software engineer in an IT company.
\end{enumerate}

\noindent\texttt{Domain of Health} 
\begin{enumerate}
    \item \textit{keyword}: lifestyle habits. \\
        \textit{Cause}: John started smoking. \\
        \textit{Effect}: Decades later, John suffers from heart disease and stroke.
    \item \textit{keyword}: health problems. \\
        \textit{Cause}: John got COVID-19. \\
        \textit{Effect}: Months later, John can still recall the bad feeling of having COVID-19.
\end{enumerate}

\noindent\texttt{Domain of Work} 
\begin{enumerate}
    \item \textit{keyword}: career success. \\
        \textit{Cause}: She has found her routine for a productive day at work. \\
        \textit{Effect}: Years later, she gets a chance for promotion because of her hardworking.
    \item \textit{keyword}: career change. \\
        \textit{Cause}: The company's new launch date puts employees under pressure. \\
        \textit{Effect}: Months later, many employees have decided to leave.
\end{enumerate}

\noindent\texttt{Domain of Business} 
\begin{enumerate}
    \item \textit{keyword}: start-up. \\
        \textit{Cause}: This newly opened coffee shop decides to attract students. \\
        \textit{Effect}: Months later, this coffee shop is the most favorite place for students to socialize.
    \item \textit{keyword}: strategy. \\
        \textit{Cause}: This company decides to expand its business overseas. \\
        \textit{Effect}: Decades later, this company is alive and well, and it is still renowned worldwide as the oldest company.
\end{enumerate}

\subsection{Annotation of Defeasibility}
\label{appendix:collection:defeasibility}

\noindent\textbf{Guideline of Annotation.\tbfspace} 
Firstly, the annotators are asked to write a supporting argument that could be the conceptual explanation behind the long-term effect. For the defeated arguments, we ask the annotators to note that with the defeater event, the effect doesn't hold any more or the effect is weakened by this defeater event. Additionally, the annotators have to specify the time interval after which the defeater happens relative to the given cause.  

\noindent\textbf{Illustrative Examples.\tbfspace} 
The illustrative examples we give for the defeasibility annotations are as follows: 
\begin{enumerate}
    \item 
    \textit{Cause}: The soccer team receives lots of funding. \\
    \textit{Effect}: Years later, the soccer team wins the soccer league championship. \\
    \textit{Supporter}: Soccer teams can hire excellent coaches and players with adequate funding.  \\
    \textit{Defeater}: Months later, the funding is wasted on corruption. 
    \item 
    \textit{Cause}: Tourists throw rubbish everywhere at the scenic spot. \\
    \textit{Effect}: Months later, fewer and fewer tourists go to this scenic spot. \\
    \textit{Supporter}: Spots with rubbish are dirty, and people don't like dirty places.  \\
    \textit{Defeater}: Volunteers pick up the trash thrown by tourists every day to keep the site clean. 
    \item 
    \textit{Cause}: John started smoking. \\
    \textit{Effect}: Decades later, John suffers from heart disease and stroke. \\
    \textit{Supporter}: The nicotine in tobacco can damage the heart.  \\
    \textit{Defeater}: Nicotine has been shown to soothe the heart.
    \item 
    \textit{Cause}: The artist signs a recording contract. \\
    \textit{Effect}: Years later, the artist becomes a star and is popular among people. \\
    \textit{Supporter}: Artists who signed contracts usually work hard to release albums.  \\
    \textit{Defeater}: The artist is lazy and rests on his laurels.    
\end{enumerate}

\section{Details of Statistics of \shorttitle} \label{appendix:statistics}
In this section, we present more details about the statistics of \shorttitle, including the sentence length distributions of supporters and defeaters~(Appendix~\ref{appendix:statistics:sentence_length}), time interval distribution~(Appendix~\ref{appendix:statistics:time_interval}), and keywords~(Appendix~\ref{appendix:statistics:keywords}). 

\subsection{Details of Sentence Length}
\label{appendix:statistics:sentence_length}
\begin{figure}[htp!]
\centering
  \centering
  \includegraphics[width=\linewidth]{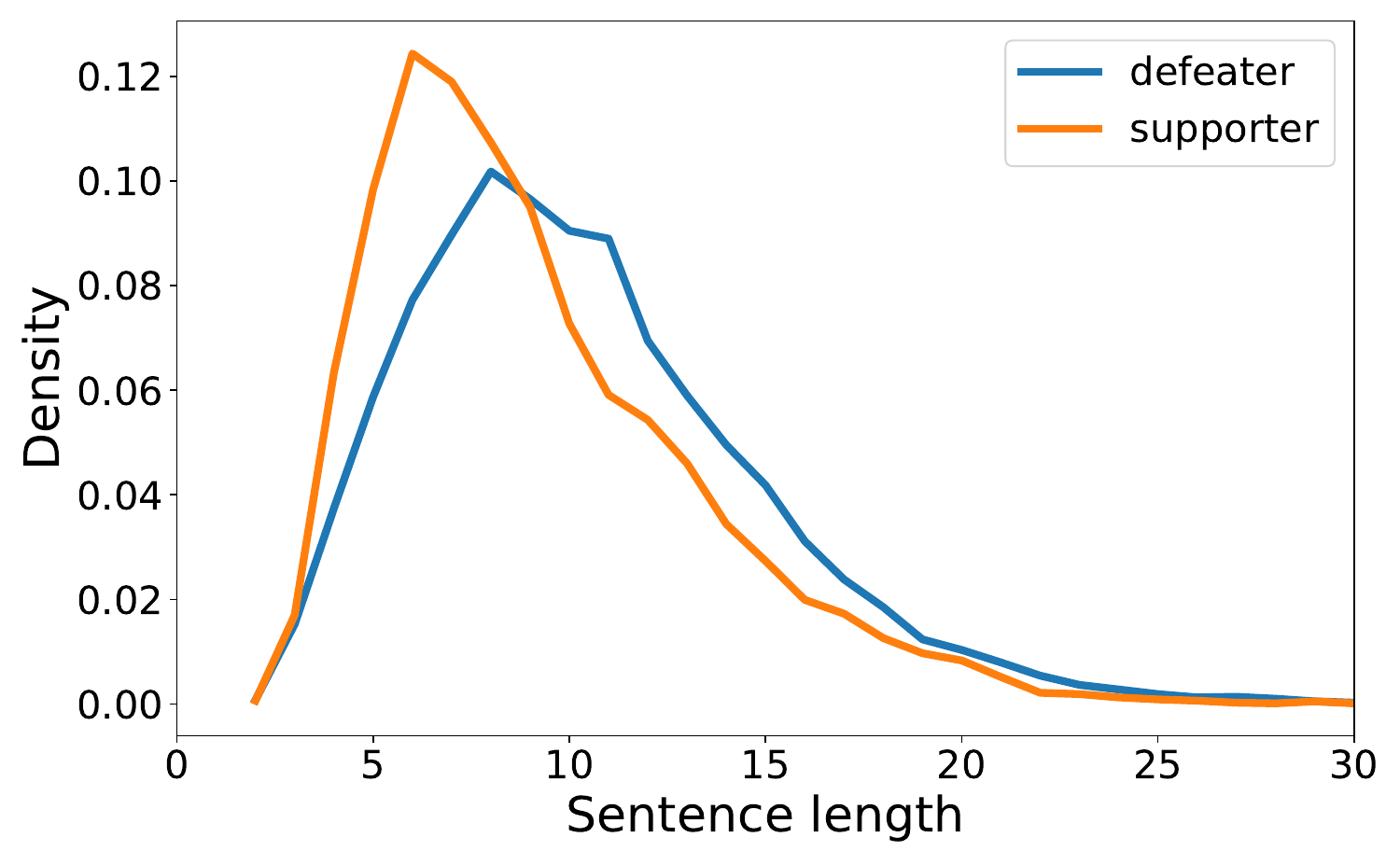}
\caption{Comparison of sentence length distributions between supporters and defeaters.}
\label{fig:appendix:sentence_length}
\end{figure}

We plot the comparison of distributions of sentence length of supporters and defeaters in Figure~\ref{fig:appendix:sentence_length}. 
Compared with supporters, defeaters are always associated with more complicated logic as they need to provide supplementary information to overturn or attenuate the causal relationship. However, supporters are relatively simple for human annotators as they only need to think out the background knowledge to provide more support for the causality relationship. 

\subsection{Details of Time Interval Distribution in Cause-Effect Pairs} \label{appendix:statistics:time_interval}
\begin{figure*}
    \centering
    \includegraphics[width=1.05\textwidth]{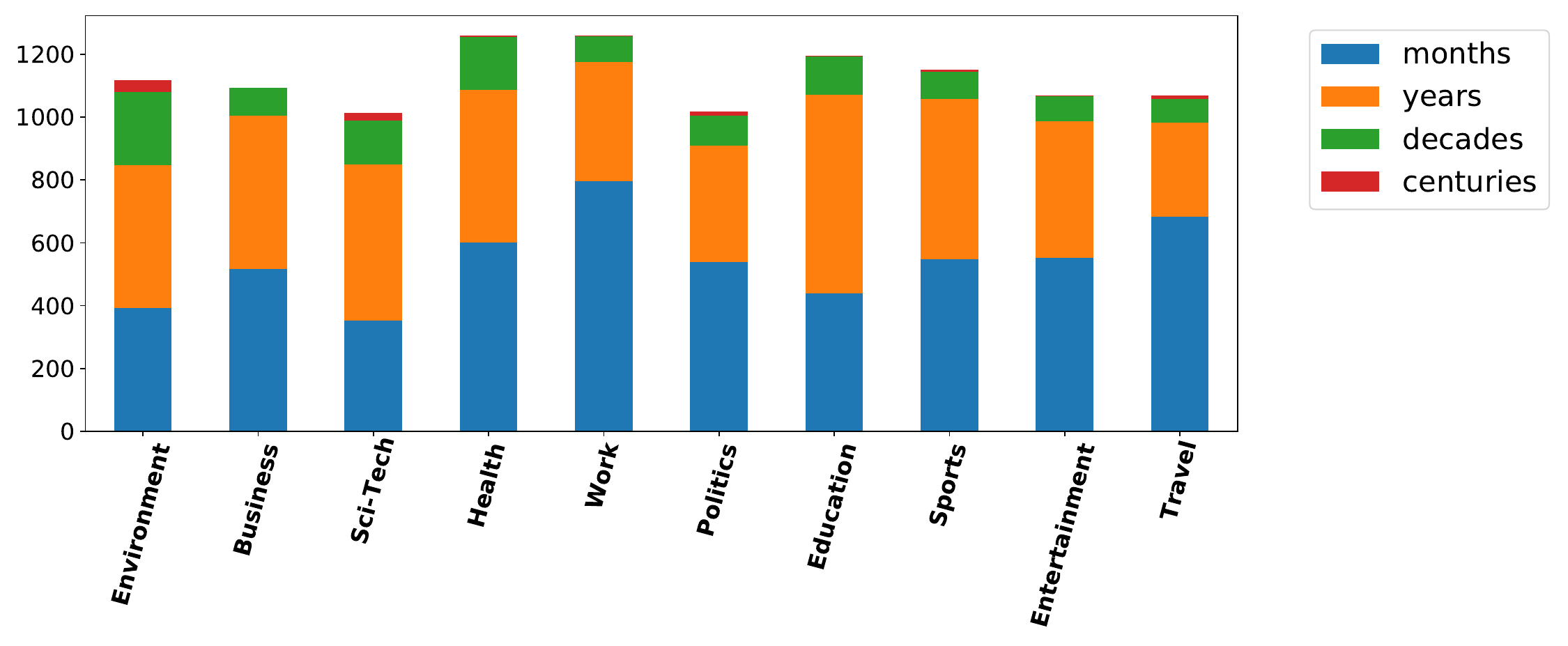}
    \caption{The distribution of time intervals in \shorttitle. We can observe that for each domain, the distribution over time label shows some variations and similarities. For instance, in the domain of environment, the portion of time labeled ``centuries later'' is relatively higher. Besides, for almost all the domains, the causalities with the time intervals of ``months later'' and ``years later'' are the most common. Besides, we could find that the causality is almost even distributed into different domains. This proves that \shorttitle is a comprehensive dataset covering different domains in daily life. }
    \label{fig:appendix:distribution}
\end{figure*}

The details of the time interval distribution in cause-effect pairs are shown in Table~\ref{tab:appendix:time} and Figure~\ref{fig:appendix:distribution}. As we can see, most time intervals fall into \texttt{months later, years later} and \texttt{decades later}. The portion of \texttt{centuries later} is relatively small. This is reasonable as it is difficult for annotators to think out some causes whose effect event happens after centuries. 
Besides, we find each domain has different time label distributions. For instance, most \texttt{centuries later} labels fall into the domain of Environment. 
\begin{table}[htb]
  \centering
   \resizebox{0.5\textwidth}{!}{  
  \begin{tabular}{l|lllll}
    \toprule
     Domains & Overall & Months & Years & Decades & Centuries \\
    \midrule
    Environment&1,118&393&455&231&39 \\
    Business&1,094&517&487&90&0 \\
    Sci-Tech&1,013&353&496&140&24 \\
    Health&1,260&601&486&168&5 \\
    Work&1,259&795&380&83&1 \\
    Politics&1,019&538&371&96&14 \\
    Education&1,195&438&634&121&2 \\
    Sports&1,151&548&509&88&6 \\
    Entertainment&1,068&553&434&80&1 \\
    Travel&1,068&682&300&75&11 \\
    \midrule
    Total & 11,245&5,418&4,552&1,172&103 \\
    \bottomrule
  \end{tabular}
  }
  \caption{Statistics of time intervals in \shorttitle.  From the statistics, we can observe that our dataset is even over different domains. Besides, we found that the number of annotations for \textit{centuries later} is relatively small, which agrees with our intuition as the effect that happens centuries later is difficult to estimate and annotate. 
  From the distribution of \shorttitle over different time intervals, we can conclude that \shorttitle is a comprehensive and unbiased dataset covering different domains and agrees quite well with the temporal characteristics in commonsense causalities of different domains. Specifically, it is more likely to obtain long-term effects in the domain of Environment, Science and Technology, and Politics than in other domains like Sports, which agrees well with human commonsense. 
  }
  \label{tab:appendix:time}
\end{table}

\subsection{Details of Keywords}
\label{appendix:statistics:keywords}
For these keywords in \S~\ref{sec:dataset:annotation}, we plot the word cloud of each domain in Table~\ref{tab:appendix:wordcloud}.  We could clearly observe that incorporating hint words into the annotation process broadens the range of topics and makes \shorttitle a comprehensive dataset that covers various aspects of commonsense knowledge. It shows the necessity of introducing keywords into the annotation of \shorttitle.

\begin{table*}
\centering
\resizebox{0.9\textwidth}{!}{
\begin{tabular}{ | p{1.8cm} | p{5cm} | p{1.7cm} | p{5cm} | }
\toprule \\
Domain & Wordcloud & Domain & Wordcloud \\ \midrule
Business & \includegraphics[width=5cm]{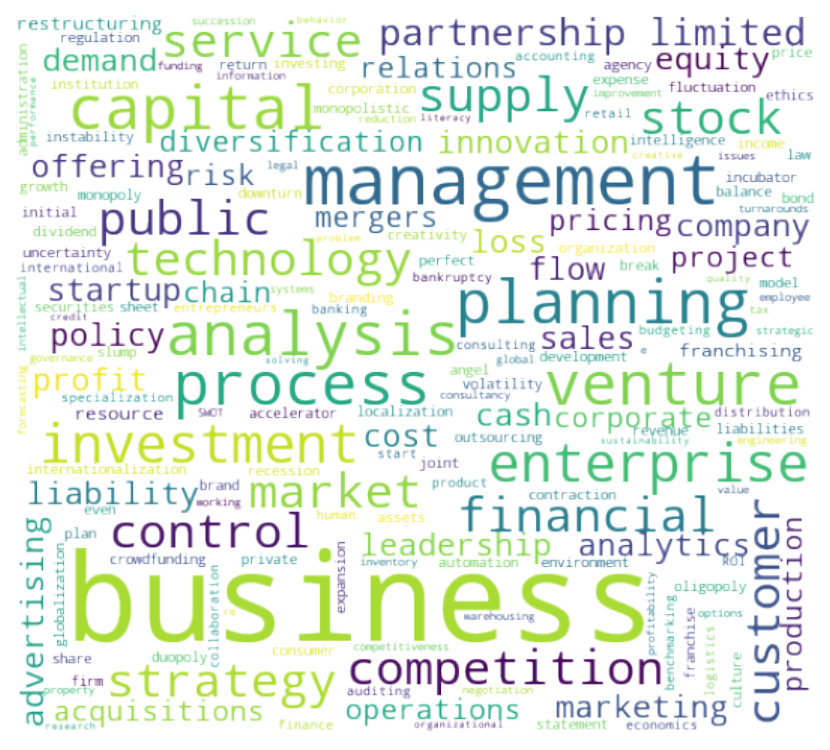} & Education & \includegraphics[width=5cm]{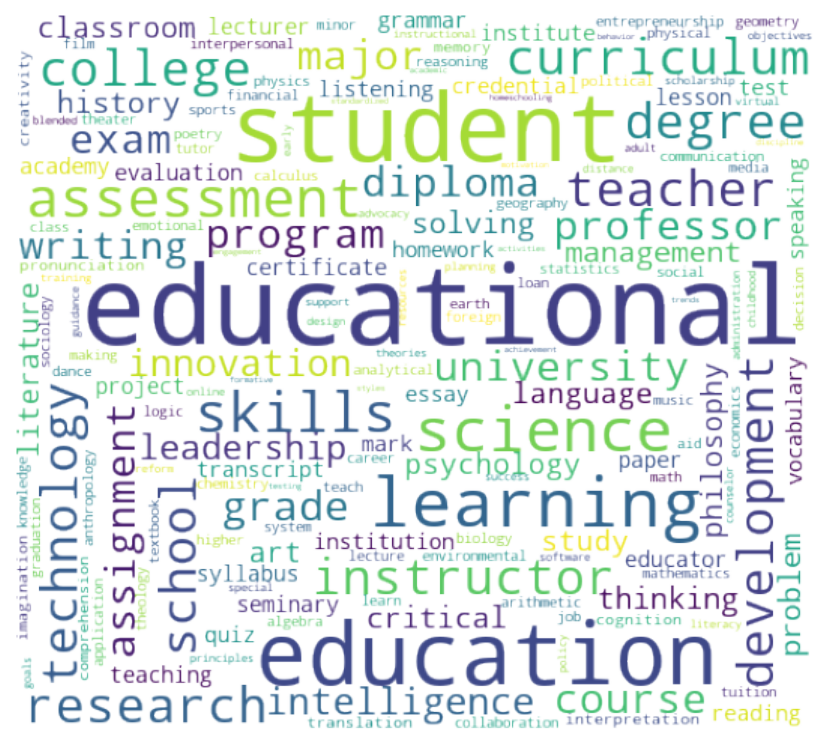} \\ \hline 
Entertainment & \includegraphics[width=5cm]{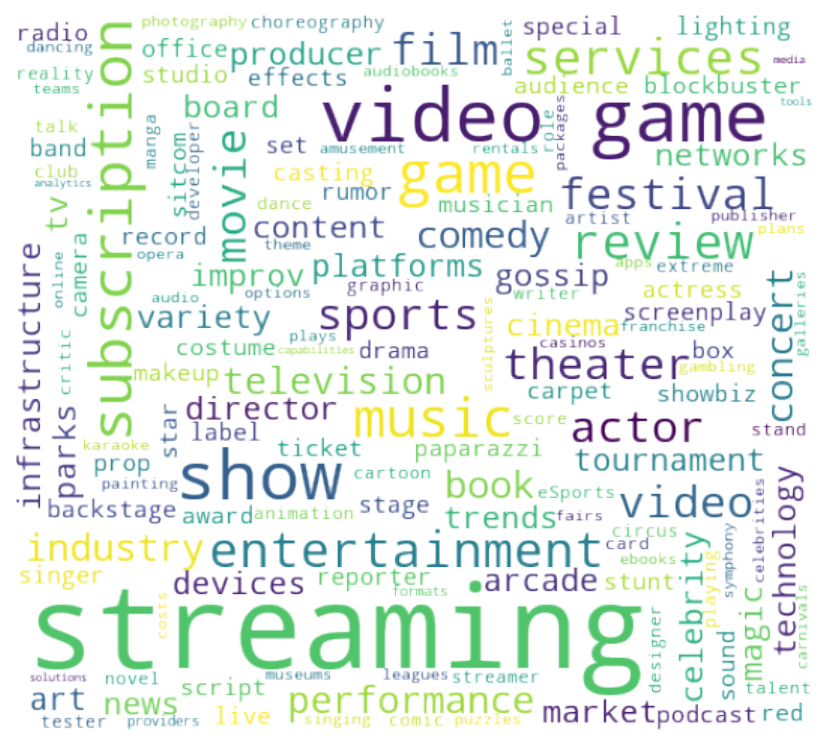} & Environment & \includegraphics[width=5cm]{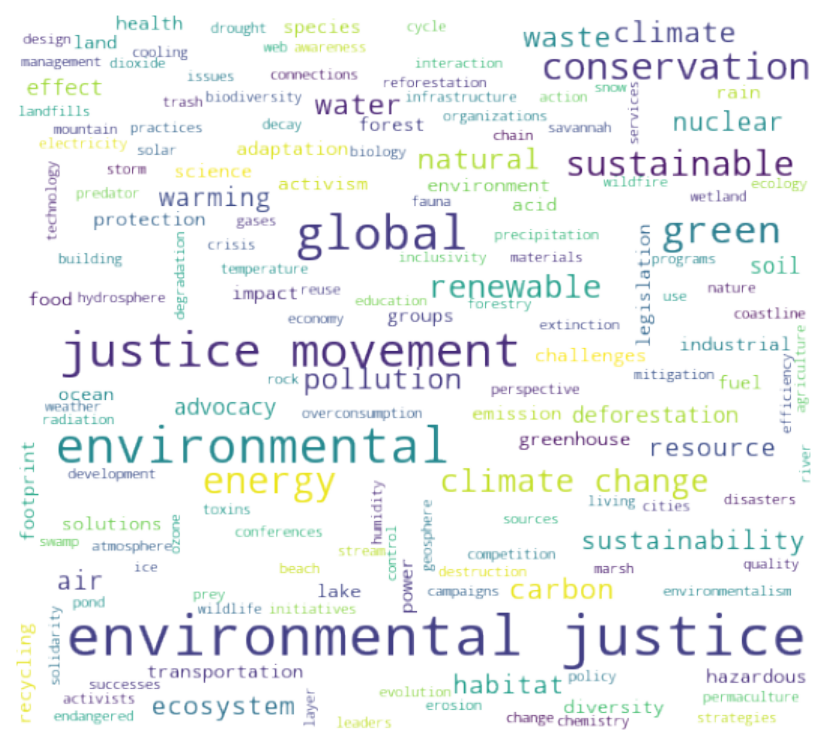} \\ \hline
Health & \includegraphics[width=5cm]{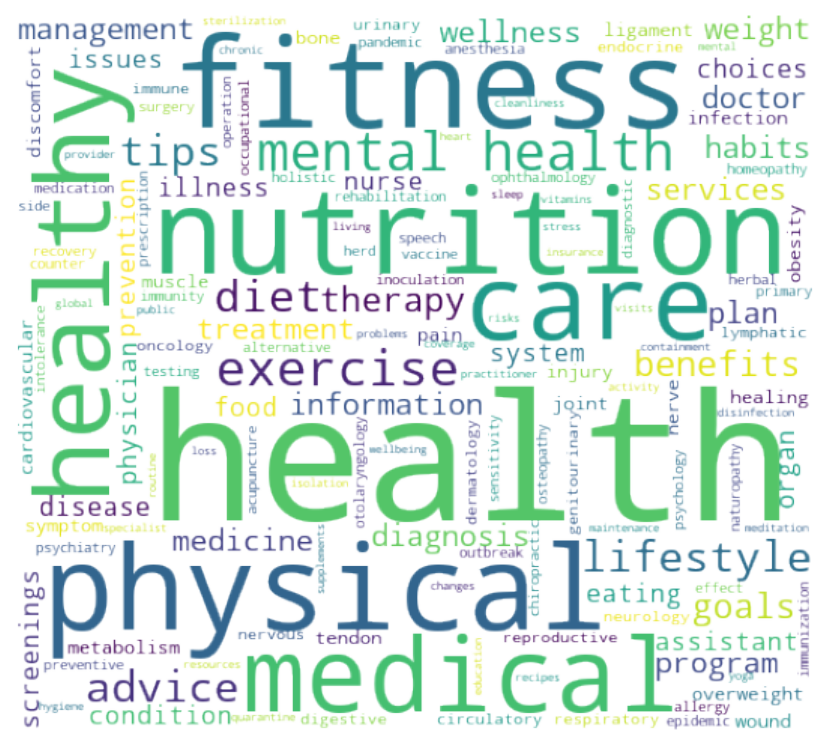} & Politics & \includegraphics[width=5cm]{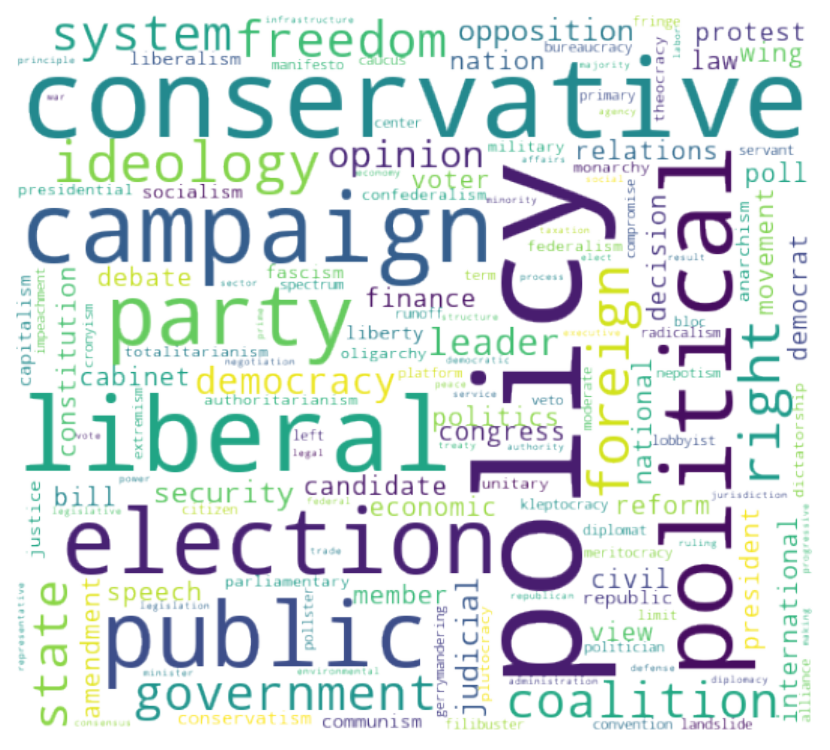} \\ \hline
Science & \includegraphics[width=5cm]{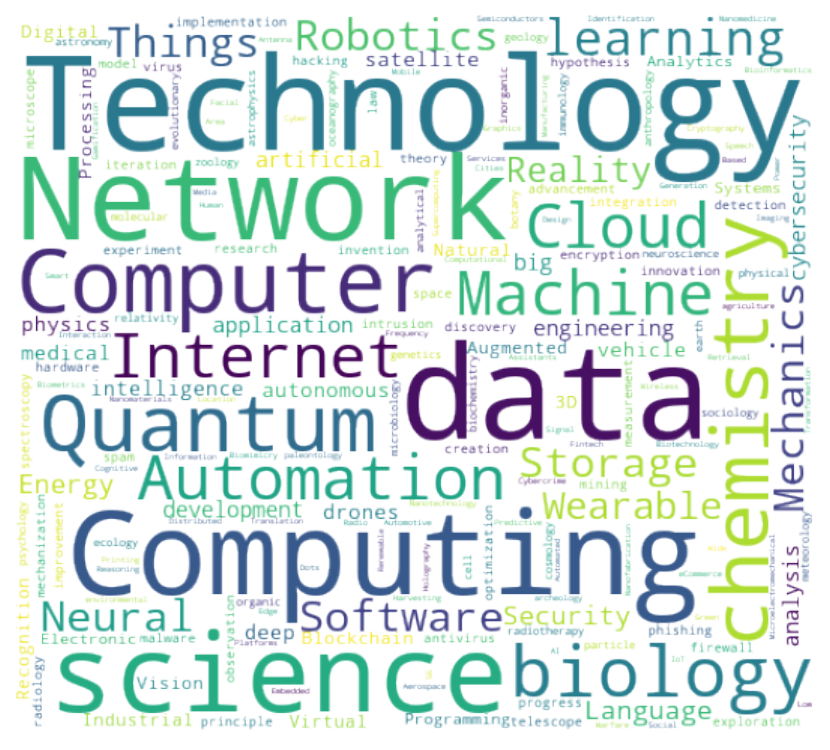} & Sports & \includegraphics[width=5cm]{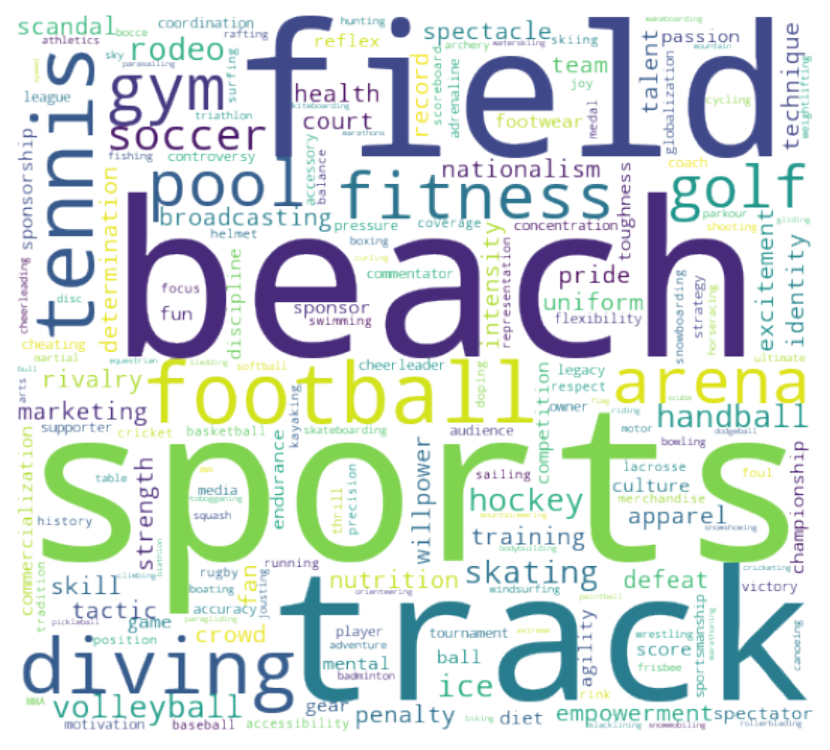} \\ \hline
Travel & \includegraphics[width=5cm]{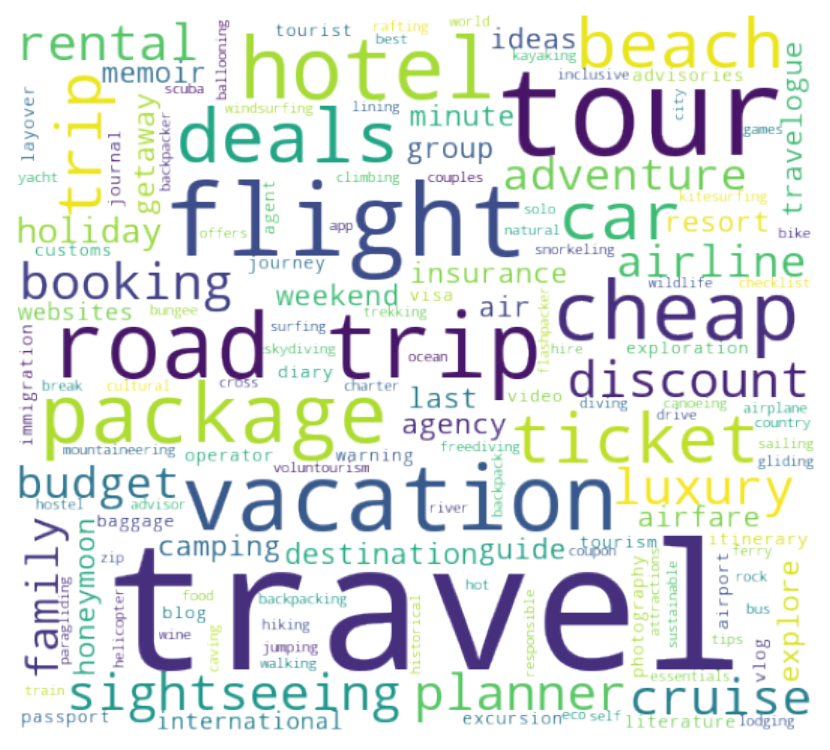} & Work & \includegraphics[width=5cm]{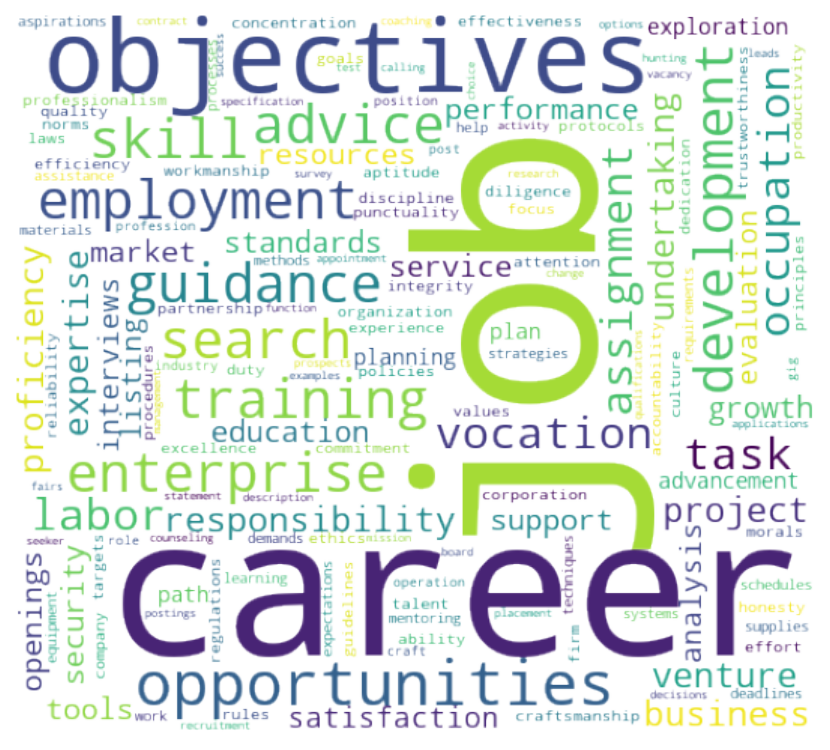} \\ \hline
\end{tabular}
}
\caption{Word cloud of each domain's keywords.}
\label{tab:appendix:wordcloud}
\end{table*}

\section{Details of Experiments on Supporter and Defeater Generation}
\label{appendix:experiments}
In this section, we give more details about the experiments of supporter and defeater generation. Specifically, we present the setup in Appendix~\ref{appendix:experiments:setup}. After that, we elaborate on the training details in Appendix~\ref{appendix:experiments:training_details}. Finally, we present the details of the human evaluation, i.e., NLP experts, in Appendix~\ref{appendix:experiments:experts}. 
\subsection{Setup} \label{appendix:experiments:setup}
In the causal defeasibility generation experiment, models are trained to generate either the supporter or the defeater given the cause and the effect (with time interval prepended). 
\begin{table}[htb]
  \begin{center}
    \begin{tabular}{p{1.5cm}p{4cm}}
      \toprule 
       Model & Input format  \\
      \midrule
      BART & <s> $C$ </s></s> $E$ </s>   \\
      T5 & $P_c$ $C$ </s> $P_l$ $E$ </s>  \\  
      GPT-2 & $P_c$ $C$ $P_l$ $E$ $P_d$  \\
      T5-large & $P_c$ $C$ </s> $P_l$ $E$ </s>  \\
      \bottomrule
    \end{tabular}
  \end{center}
  \caption{Input format of each model in causal defeasibility generation task. $P_c=\text{"cause:"}$, $P_l=\text{"long-term effect:"}$, $E$ is the effect prepended with the long time interval. For example, $E_l$ could be "Months later, mental health issues arise.". $C$ is the cause of the effect. These models have different suggestions for input format. To achieve the best performance, we follow the official suggestions in their original papers. This is the reason why we use different input formats for different models. }
  \label{tab:appendix:input_format:defeasibility}
\end{table}
For sequence-to-sequence models, the encoder input is to use the text prefixes $P_c$ and $P_l$ for the cause and long-term effect. For the causal language model GPT-2, we use the text prefix $P_d$ for defeasibility, which could be either "assumption:" or "defeater:" depending on the task.
The input format for all models involved is shown in Table~\ref{tab:appendix:input_format:defeasibility}. 
The versions of all the packages, tools~(scientific artifacts) are listed in the \texttt{requirements.txt}, which is included in our code repository. The code repository is attached as the supplementary zip file. 
The major tools we use include \texttt{PyTorch}~\footnote{https://pytorch.org/}~(BSD-3 License) and \texttt{transformers}~\footnote{https://huggingface.co/docs/transformers/index}~(Apache License 2.0). 
Our proposed dataset \shorttitle~(MIT License) will be released to the public after this paper's acceptance. Besides, the other datasets we use, e-CARE~(MIT License)~\cite{du2022care} and COPA~(BSD 2-Clause License)~\cite{roemmele2011choice}, are both publicly available and open-source. 

\subsection{Training Details} \label{appendix:experiments:training_details}
We fine-tune BART-base (140M), GPT-2 (117M), T5-base (223M), and T5-large (783M) models. We use the Huggingface Trainer default optimizer AdamW for training. All models are optimized with a batch size of 32 (8 for each GPU) and finetuned for 3 epochs. The learning rate is set to 1e-5 for the BART and GPT-2 models and 3e-4 for the T5-base and T5-large models.
The experiments of generation on \shorttitle are conducted by a single run. The prompt experiments for keyword generation and CTCW are conducted with different prompts to search for the best prompts. 
All of the experiments are run with a machine with four GPUs. All of these GPUs are NVIDIA TITAN X (Pascal) with a memory size of 12,288MB. 
The random seeds for the single run experiments in both defeater/supporter generation and CESAR are set as 42. 
\subsection{Details of NLP Experts in Our Evaluation} \label{appendix:experiments:experts}
All of these three NLP experts are graduate students whose major research interests are natural language processing. All of them have taken advanced NLP courses and have relative research experience and two of them have papers in the domain of NLP. 

\section{Details of Existing Causal Strength Metrics}
\label{appendix:related_metrics}
In this section, we mainly introduce the details of existing metrics on causal strength. 

\noindent\textbf{Causal Explanation Quality~(CEQ) Score.\tbfspace} 
The causal strength in CEQ ~\cite{du2022care,luo-etal-2016-commonsense} is defined as: 
\begin{equation}
    \causalstrength{C}{E} = \frac{1}{N_{C} + N_{E}} \sum_{w_i \in C, w_j \in E} \text{cs}(w_i, w_j) 
\end{equation}
where $N_C$ and $N_E$ are the number of words in $C$ and $E$. $\text{cs}(w_i, w_j)$ is the causal strength between these two words: $\text{cs}(w_i, w_j) = \frac{\text{Count}(w_i, w_j)}{\text{Count}(w_i) \text{Count}^{\alpha}(w_j)}$, where $w_i$ is from the cause event while $w_j$ is from the effect event. The term $\text{Count}(w_i, w_j)$ denotes the frequency with which $w_i$ and $w_j$ co-occur in causal statements, while $\text{Count}(w_i)$ indicates the total number of appearances in such sentences.

\noindent\textbf{ROCK.\tbfspace} 
ROCK~\cite{zhang2022rock} defines a causal strength metric from the perspective of causal inference~\cite{hernan2010causal}: 
\begin{equation}
\resizebox{0.85\hsize}{!}{
$
\begin{aligned}
    \causalstrength{C}{E} &= \mathbb{E}_{x} [\mathbb{P}(C \prec E \vert x ) - \mathbb{P}(\neg C \prec E \vert x)] \\
    & \approx f(C, E) - \frac{1}{\vert \mathcal{A}' \vert} \sum_{A \in \mathcal{A}'} f(A, E)
\end{aligned} 
$
}
\end{equation}
where $x$ is the confounder of the cause event and the effect event. $\neg C$ is the intervention of the cause $C$. $f(C,E)$ is an estimate for $\mathbb{P}(C \prec E)$, i.e., the probability of $C$ happens before $E$. $\mathcal{A}'$ is the $L_p$-constrained set for the generated interventions conditioned on confounders: $\mathcal{A}' \coloneqq \{ A \in \mathcal{A}: \frac{1}{\vert \mathcal{X} \vert} \Vert q(x; A) - q(x, C)\Vert_{2} \leq \epsilon \}$,
where the set $\mathcal{A}$ contains generated interventions $\neg C$ and $\mathcal{X}$ is the set of the generated confounders $x$. $\epsilon$ is the threshold. $q(x; E)$ is the temporal propensity measuring the conditional probability of the event $E$ given an event $x$.

\section{More Details of CESAR} \label{appendix:cesar}
In this section, we present more details of CESAR. Specifically, we present its setup in Appendix~\ref{appendix:cesar:setup}, its score computation pipeline in Appendix~\ref{appendix:cesar:score}, preparation of CESAR's training data in Appendix~\ref{appendix:cesar:training_data}, discussion of the concatenation operation in Appendix~\ref{appendix:cesar:concatenation}, and ablation study in Appendix~\ref{appendix:cesar:ablation}. 

\subsection{Setup of CESAR} \label{appendix:cesar:setup}
The CESAR model consists of a BERT~\cite{devlin2019bert} model reinforced with the causality-aware attention layer. We utilize the pre-trained \texttt{bert-large-uncased} model from Hugging Face~\cite{wolf-etal-2020-transformers}. This BERT model has an embedding dimension $d$ of 1024. Thus, causality-aware attention has two learnable weight matrices $\mathbf{W}_q, \mathbf{W}_k \in \mathbb{R}^{1024 \times 1024}$. The input to the model is the output of the respective tokenizer for the \texttt{bert-large-uncased} model. Specifically, we jointly preprocess the given cause $C$ and the given effect $E$ with the tokenizer, which produces the respective \textit{input\_ids} with appropriately added special tokens. \textit{token\_type\_ids} of $C$ are marked with 0 and \textit{token\_type\_ids} of $E$ are marked with 1. \textit{attention\_mask} is also included so that the model avoids attending on [PAD] tokens. Hence, the model input consists of \textit{input\_ids}, \textit{token\_type\_ids}, and \textit{attention\_mask}. The maximal input size for the CESAR model is 512 tokens including [CLS] and [SEP] which are appended at the beginning and the end of the sequence, respectively. 

\subsection{Score Computation} \label{appendix:cesar:score}
The CESAR score is computed in several stages. 
\begin{enumerate}
    \item we extract the token embeddings from the BERT's last hidden layer:
\begin{align*}
    (\mathbf{C} + \mathbf{E}) = \text{BERT}(&\text{input\_ids}, \\ 
    &\text{token\_type\_ids}, \\ &\text{attention\_mask}).
\end{align*}
Since we jointly preprocess cause and effect, BERT also jointly produces embeddings for tokens of the cause and effect. Hence, $(\mathbf{C} + \mathbf{E})\in \mathbb{R}^{(n+m) \times d}$, where $n$ is the length of the cause while $m$ is the length of the effect. In this setting, \textit{token\_type\_ids} suggests to BERT which tokens belong to cause and which belong to effect.
\item Based on \textit{token\_type\_ids}  and \textit{attention\_mask}, model separates embeddings to $\mathbf{C} \in \mathbb{R}^{n \times d}$ and $\mathbf{E} \in \mathbb{R}^{m \times d}$. Next, embeddings are given to the causality-aware attention layer, where we obtain query $\mathbf{Q} = \mathbf{C}\mathbf{W}_q$ and key $\mathbf{K} = \mathbf{E}\mathbf{W}_k$ vectors. These vectors give us the attention scores for token pairs as $\mathbf{A} = \text{softmax}\left(\mathbf{QK}^T\right)$ where softmax is performed over all values of the matrix (not only over a single dimension as in the conventional attention layers).
\item The causal strength value given by CESAR is calculated as follows, 
\begin{align}
      \causalstrength{C}{E} &= \sum_{i=1}^{n} \sum_{j=1}^{m} a_{ij}\frac{|c_i^T e_j|}{\Vert c_i\rVert\Vert e_j\rVert}
\end{align}
where $c_i\in\mathbf{C}$ and $e_j\in\mathbf{E}$ represent causal embeddings of tokens of $C$ and $E$ respectively, and $a_{ij}\in\mathbf{A}$ is the attention weights put on each pair of tokens. Please note that we keep [CLS] and [SEP] special tokens when computing the causal strength with CESAR. As a result, the first token of cause representation is always a [CLS] token. See Section~\ref{appendix:cesar:ablation} for more details about the roles of these two special tokens.
\end{enumerate}

\subsection{Preparation for Training Data} \label{appendix:cesar:training_data}
\noindent\textbf{Training Data. \tbfspace}
We train the CESAR metrics on the augmented e-CARE dataset \cite{du2022care}. We consider both the dev and train parts of e-CARE and combine them into a single dataset. This dataset contains causal-effect sentence pairs with a conceptual explanation for each cause-effect pair. Accordingly, with the conceptual explanation, the causal strength between the cause and the effect increases. 
Motivated by this fact, we set the causal strength as $\causalstrength{C}{E} = 0.7$ and $\causalstrength{C \oplus H}{E} = 1.0$ where $H$ is the conceptual explanation for why $C$ leads to the occurrence of $E$. The dataset also includes pairs of sentences with no causal relationship, we set the causal strength of these non-causal event pairs to $0.0$. Lastly, in order to replicate the decreasing effect that defeater has on causal strength as proposed in the \shorttitle, we use ChatGPT to generate semantic opposites from the conceptual explanations provided in e-CARE and set $\causalstrength{C \oplus \neg H}{E} = 0.2$ where $ \neg H$ is an opposite of the conceptual explanation for $C$ and $E$. 
$\neg H$ is generated by ChatGPT and the prompt given to ChatGPT to generate this semantic opposite is shown in the next paragraph. The training dataset constructed in this manner consists of a total of 68,220 examples.
The discussion about the impacts of these various types of data samples is described in Appendix~\ref{appendix:cesar:ablation}.

\noindent\textbf{Data Augmentation with ChatGPT. \tbfspace} As described in the aforementioned part, in order to imitate the decreasing effect of the defeaters in \shorttitle, we use ChatGPT to generate sentences that have opposite meanings from the conceptual explanation provided in the e-CARE dataset. To be more specific, we use OpenAI's API~\footnote{https://openai.com/blog/chatgpt} and the gpt-3.5-turbo model with a temperature set to 0.9. The prompt provided to the gpt-3.5-turbo model stands as follows:

\begin{center}
\mybox[gray!20]{
You are a helpful assistant that helps to find the opposite of the given sentence. The real truth is not important just the resulting sentence must be of the opposite meaning, negating the information that the given sentence tries to convey. Try to not give a simple negation. Output ONLY the resulting sentence, nothing else. For example for the prompt: "Friends join communities.", the output should be: "Friends avoid communities." Also for the prompt: "Sulfonamides cause hemolysis less commonly.", the output should be "Sulfonamides cause hemolysis more commonly.". Another example would be that for the prompt: "Homelessness greatly increases the likelihood of a suicide attempt.", the output is: "Homelessness greatly decreases the likelihood of a suicide attempt." The last example is that for the prompt: "Production occurs in dense regions.", the output must be: "Production occurs only in sparse regions.
\\
\\
"Products cause slow growth."
}
\end{center}
The resulting response to the above prompt is: "Products promote rapid growth". For most cases, we observe that the model generates the satisfying negation. However, there are examples where it applies the double negation such that the second negation nullifies the first thus resulting sentence is not semantically opposite to the initial input. For instance, for the prompt "Attempts yield results", the output is "Not attempting ensures no results", or for "The sun sets early in December", we get "The sun rises late in December". We believe that doubly negated sentences incorporated in the training set and generated as defeaters, but not effectively acting as such, introduce extraneous noise, thereby impeding the model's performance in identifying defeaters. Accordingly, our model demonstrated superior results in supporters as compared to defeaters. In addition, despite an instruction to avoid simplistic negation by the mere introduction of "not" in input sentences, the gpt-3.5-turbo model continues to do so in a considerable number of examples. All of this leads us to the conclusion that we could improve the performance of our model if we would generate the opposites of conceptual explanations from e-CARE more reliably and correctly, e.g., by using human labor instead of automatic rendering.

\subsection{Formulation of Concatenation Operation} \label{appendix:cesar:concatenation}
To validate the expected behavior of CESAR, it is necessary to demonstrate the capacity of CESAR to capture the causal strength changes after integrating supporters and defeaters with respective causes. In order to do that, we need to first formulate the concatenation operation $\oplus$ between two statements. Since we use the BERT model as our backbone model when implementing CESAR, we define concatenation as follows
\begin{equation}
    C \oplus A/D = C \hspace{1.5mm} \text{[SEP]}\hspace{1.5mm} A/D
\end{equation}
where $C$ is the cause that we wish to concatenate with either supporter $A$ or defeater $D$, and [SEP] is BERT's special token that helps BERT know that $C$, $A/D$, and $E$ are separate sentences~(entities).

\subsection{Ablation  Study} \label{appendix:cesar:ablation}
\begin{table}[htb!]
    \centering
    \resizebox{0.5\textwidth}{!}{
    \begin{tabular}{p{5.5cm}p{1.5cm}p{1cm}p{1.5cm}}
      \toprule 
       & Supporter & Defeater & Geometric mean\\
      \midrule
      CESAR & 84.6 & 75.8 &  \textbf{80.1} \\
      \midrule
      w/o causality-aware attention & 91.2 &  64.4 & 76.6\\
      w/o [CLS] \& [SEP] &  80.2 & 76.0 & 78.0 \\
      w/o data augmentation &  64.2 & 63.6 & 63.9 \\
      w/ imbalanced data augmentation &  89.0 & 24.4 & 46.6 \\
      w/ \texttt{bert-large-cased} &  76.8 & 78.0 & 77.9 \\
      w/ \texttt{bert-base-uncased} &  78.6 & 79.8 & 79.2\\
      \bottomrule
    \end{tabular}
  }
  \caption{{Performance on 500 samples from \shorttitle by different variations of the CESAR metrics. 
  The accuracy on supporters and defeaters is calculated in the same ways as that described in Table~\ref{tab:causal_metrics}. 
  For more clarity, we employ the abbreviation "w/o" to indicate the exclusion of a specific component from the CESAR build-up. Therefore, we conduct experiments by removing the causality-aware attention layer, as well as the [CLS] and [SEP] tokens. Furthermore, we train CESAR without using the augmented dataset containing supporters and defeaters. Conversely, we employ the abbreviation "w/" to denote that a particular component has been substituted from the original CESAR build-up. 
  Besides, we explore imbalanced data augmentation, where only conceptual explanations are augmented as supporters, without generating their opposites. Finally, we evaluate the usage of alternative BERT models including \texttt{bert-large-cased} and \texttt{bert-base-uncased} instead of the original \texttt{bert-large-uncased}.}
  }
  \label{tab:appendix:causal_metrics_ablation}
\end{table}
There are many techniques contributing to the success of CESAR in capturing the causal strength changes such as causality-aware attention, special tokens like [CLS] and [SEP], data augmentation, and backbone model selection. 
To validate the effectiveness of these techniques, we conducted a comprehensive ablation study. In Table~\ref{tab:appendix:causal_metrics_ablation}, we display the results with various ablations from the best CESAR model. From the results, we have the following observations: 
\newcommand{\ablationfont}[1]{{#1}}
\begin{itemize}
    \item \ablationfont{w/o causality-aware attention}: our results demonstrate the pivotal contribution of the causality-aware attention layer in enhancing metrics stability and performance in scenarios involving defeating information. Notably, with the incorporation of causality-aware attention, we observe a substantial improvement in accuracy---from 64.4\% to 75.8\%---on defeaters. Specifically, this layer enables redirection of focus~(attention) from word pairs with strong causal relationships to those with weaker associations following the introduction of the defeaters. One interesting phenomenon here is that the ablated version of CESAR in this setting, i.e., \textit{w/o causality-aware attention}, achieves an accuracy of 91.2\% in capturing the causal strength change brought by supporters, which is even better than CESAR. However, this ablated version struggles in capturing the causal strength changes with defeaters, with an accuracy of 64.4\%. In other words, the causality-aware attention mechanism makes CESAR a more comprehensive evaluation metric on causal strength, which can not only capture the supplementary information that increases the causal strength but also can capture the counterpart that decrease the causal strength.  
    \item \ablationfont{w/o [CLS] \& [SEP]}: we observe highly unstable training once we attempt to remove [CLS] and [PAD] tokens when computing the causal strength score. Specifically, the loss during training exhibits high fluctuations with our default learning rate of $1e-5$. If the learning rate is decreased, the optimizer has trouble finding a satisfying local minimum, and training is slow. Hence, we incorporate [CLS] and [PAD] tokens when calculating the causal strength. The introduction of these special tokens is due to considerations for training stability. Also, using these tokens can also enhance the performance a bit: from 80.2\% to 84.6\% on supporters.  
    
    \item \ablationfont{w/o data augmentation}: 
    there are four kinds of data samples for CESAR's training: 
    (a) True cause-effect pairs with a causal strength value of 0.7, i.e., $\causalstrength{C}{E} = 0.7$. 
    (b) False cause-effect pairs that do not have a causal relationship with a causal strength value of 0.0. 
    (c) cause-explanation-effect triples with a causal strength value of 1.0, i.e., $\causalstrength{C \oplus H}{E}$ to 1.0 where $H$ is the explanation for the causal relationship between $C$ and $E$. 
    (d) cause-opposite\_explanation-effect triples with a causal strength value of 0.2, i.e., $\causalstrength{C \oplus \neg H}{E}$ to 0.2 where $\neg H$ is an opposite of the conceptual explanation for $C$ and $E$. $\neg H$ is generated by ChatGPT. For the ablation case \textit{w/o data augmentation}, we only use the data samples of type (a) and (b). We can clearly notice that data augmentation plays a crucial role as the accuracy decreases from 84.6\% to 64.2\% in capturing the causal strength changes brought by supporters, and from 75.8\% to 63.6\% on defeaters. It shows the data augmentation with explanation and its opposite is a necessary component for the success of CESAR. It can be explained that the CESAR is provided with more fine-grained examples in understanding different levels of intensity of causal strength. 
    
    \item \ablationfont{w/ imbalanced data augmentation}: for the ablated case \textit{w/ imbalanced data augmentation}, we only use data samples of type (a), (b), and (c). We observe that without the generated opposites of explanations $\neg H$, CESAR becomes overly biased as it seems to learn to always increase causal strength once the new information is attached to the cause. The accuracy on defeaters decreases from 75.8\% to 24.4\%. It proves that the opposites of conceptual explanations play a critical role in CESAR.  
    
    \item \ablationfont{w/ \texttt{bert-large-cased}}:  we experiment with a BERT variant that distinguishes cased and uncased words, which decreases the overall performance, presumably due to heightened complexity and little value added to the metric. Note that our CESAR is built upon a \texttt{bert-large-uncased} model. 
    
    \item \ablationfont{w/ \texttt{bert-base-uncased}}: we assess the efficacy of CESAR based on alternative backbone models. As one can observe in Table~\ref{tab:appendix:causal_metrics_ablation}, it indicates that the employment of \texttt{bert-base-uncased} yields comparable results to its larger counterpart, \texttt{bert-large-uncased}. Strikingly, the former option, i.e.,  \texttt{bert-base-uncased}, attains the best defeater score compared to all other configurations, thereby suggesting its utility as a viable alternative in resource-constrained settings. 
\end{itemize}

\end{document}